\documentclass[journal,comsoc]{IEEEtran}
\usepackage[T1]{fontenc}% optional T1 font encoding

\ifCLASSINFOpdf
\else
\fi
\usepackage{amsmath}
\usepackage[cmintegrals]{newtxmath}
\usepackage{url}
\usepackage{times}
\usepackage{epsfig}
\usepackage{graphicx}
\usepackage{amsmath}

\usepackage[utf8]{inputenc} % allow utf-8 input
\usepackage[T1]{fontenc}    % use 8-bit T1 fonts
\usepackage{booktabs}       % professional-quality tables
\usepackage{amsfonts}       % blackboard math symbols
\usepackage{nicefrac}       % compact symbols for 1/2, etc.
\usepackage{microtype}      % microtypography
\usepackage{algorithm}
\usepackage{algpseudocode}
\usepackage{wrapfig}
\usepackage{pdfpages}
\usepackage{array}
\usepackage{bm}
\usepackage{multirow}
\usepackage{enumitem}
\usepackage{makecell}
\usepackage{color}

\usepackage{caption}
\usepackage{subcaption}

\def\E{{\rm E}}

\def\KL{{\rm KL}}

\begin{document}
%
% paper title
% Titles are generally capitalized except for words such as a, an, and, as,
% at, but, by, for, in, nor, of, on, or, the, to and up, which are usually
% not capitalized unless they are the first or last word of the title.
% Linebreaks \\ can be used within to get better formatting as desired.
% Do not put math or special symbols in the title.
\title{Energy-Based Continuous Inverse Optimal Control}
%
%
% author names and IEEE memberships
% note positions of commas and nonbreaking spaces ( ~ ) LaTeX will not break
% a structure at a ~ so this keeps an author's name from being broken across
% two lines.
% use \thanks{} to gain access to the first footnote area
% a separate \thanks must be used for each paragraph as LaTeX2e's \thanks
% was not built to handle multiple paragraphs
%

\author{Yifei Xu, Jianwen Xie, Tianyang Zhao, Chris Baker, Yibiao Zhao and~Ying Nian Wu % <-this % stops a space
\IEEEcompsocitemizethanks{
\IEEEcompsocthanksitem Y. Xu is with the Department of Statistics, University of California, Los Angeles, CA 90095, USA. E-mail: fei960922@ucla.edu 
\IEEEcompsocthanksitem J. Xie is with the Cognitive Computing Lab, Baidu Research, Bellevue, WA 98004, USA. E-mail: jianwen@ucla.edu
\IEEEcompsocthanksitem T. Zhao is with the Department of Statistics, University of California, Los Angeles, CA 90095, USA. E-mail: tyzhao@ucla.edu 
\IEEEcompsocthanksitem C. Baker is with iSee Inc., Cambridge, MA 02139, USA. E-mail: chrisbaker@isee.ai
\IEEEcompsocthanksitem Y. Zhao is with iSee Inc., Cambridge, MA 02139, USA. E-mail: yz@isee.ai
\IEEEcompsocthanksitem Y. N. Wu is with the Department of Statistics, University of California, Los Angeles, CA 90095, USA. E-mail: ywu@stat.ucla.edu 
% note need leading \protect in front of \\ to get a newline within \thanks as
% \\ is fragile and will error, could use \hfil\break instead.
}% <-this % stops an unwanted space
}

% note the % following the last \IEEEmembership and also \thanks - 
% these prevent an unwanted space from occurring between the last author name
% and the end of the author line. i.e., if you had this:
% 
% \author{....lastname \thanks{...} \thanks{...} }
%                     ^------------^------------^----Do not want these spaces!
%
% a space would be appended to the last name and could cause every name on that
% line to be shifted left slightly. This is one of those "LaTeX things". For
% instance, "\textbf{A} \textbf{B}" will typeset as "A B" not "AB". To get
% "AB" then you have to do: "\textbf{A}\textbf{B}"
% \thanks is no different in this regard, so shield the last } of each \thanks
% that ends a line with a % and do not let a space in before the next \thanks.
% Spaces after \IEEEmembership other than the last one are OK (and needed) as
% you are supposed to have spaces between the names. For what it is worth,
% this is a minor point as most people would not even notice if the said evil
% space somehow managed to creep in.

% The paper headers
\markboth{ }%
{Shell \MakeLowercase{\textit{et al.}}: Bare Demo of IEEEtran.cls for IEEE Communications Society Journals}
% The only time the second header will appear is for the odd numbered pages
% after the title page when using the twoside option.
% 
% *** Note that you probably will NOT want to include the author's ***
% *** name in the headers of peer review papers.                   ***
% You can use \ifCLASSOPTIONpeerreview for conditional compilation here if
% you desire.

% If you want to put a publisher's ID mark on the page you can do it like
% this:
%\IEEEpubid{0000--0000/00\$00.00~\copyright~2015 IEEE}
% Remember, if you use this you must call \IEEEpubidadjcol in the second
% column for its text to clear the IEEEpubid mark.

% use for special paper notices
%\IEEEspecialpapernotice{(Invited Paper)}

% make the title area
\maketitle

\begin{abstract}
The problem of continuous inverse optimal control (over finite time horizon) is to learn the unknown cost function over the sequence of continuous control variables from expert demonstrations. In this article, we study this fundamental problem in the framework of energy-based model, where the observed expert trajectories are assumed to be random samples from a probability density function defined as the exponential of the negative cost function up to a normalizing constant. The parameters of the cost function are learned by maximum likelihood via an ``analysis by synthesis'' scheme, which iterates (1) synthesis step: sample the synthesized trajectories from the current probability density using the Langevin dynamics via back-propagation through time, and (2) analysis step: update the model parameters based on the statistical difference between the synthesized trajectories and the observed trajectories. Given the fact that an efficient optimization algorithm is usually available for an optimal control problem, we also consider a convenient approximation of the above learning method, where we replace the sampling in the synthesis step by optimization. Moreover, to make the sampling or optimization more efficient, we propose to train the energy-based model simultaneously with a top-down trajectory generator via cooperative learning, where the trajectory generator is used to fast initialize the synthesis step of the energy-based model. We demonstrate the proposed methods on autonomous driving tasks, and show that they can learn suitable cost functions for optimal control.

\end{abstract}

% Note that keywords are not normally used for peerreview papers.
\begin{IEEEkeywords}
Inverse optimal control; Energy-based models; Langevin dynamics; Cooperative learning.
\end{IEEEkeywords}

\IEEEpeerreviewmaketitle

\section{Introduction}\label{sec:introduction}

\subsection{Background and motivation}

\IEEEPARstart{T}{he} problem of continuous optimal control has been extensively studied. In this paper, we study the control problem of finite time horizon, where the trajectory is over a finite period of time. In particular, we focus on the problem of autonomous driving as a concrete example. In continuous optimal control, the control variables or actions are continuous. The dynamics is known. The cost function is defined on the trajectory and is usually in the form of the sum of stepwise costs and the cost of the final state. We call such a cost function Markovian. The continuous optimal control seeks to minimize the cost function over the sequence of continuous control variables or actions, and many efficient algorithms have been developed for various optimal control problems \cite{todorov2006optimal}. For instance, in autonomous driving, the iLQR (iterative linear quadratic regulator) algorithm is a commonly used optimization algorithm \cite{li2004iterative,bemporad2002explicit}. We call such an algorithm the built-in optimization algorithm for the corresponding control~problem.

	In applications such as autonomous driving, the dynamics is well defined by the underlying physics and mechanics. However, it is a much harder problem to design or specify the cost function. One solution to this problem is to learn the cost function from expert demonstrations by observing their sequences of actions. Learning the cost function in this way is called continuous inverse optimal control (IOC) problem. 
	
	In this article, we study the fundamental problem of continuous inverse optimal control in the framework of energy-based model \cite{xie2016theory}. Originated from statistical physics, an energy-based model (EBM) is a probability distribution where the probability density function is in the form of exponential of the negative energy function up to a normalizing constant. The energy function maps the input into a scalar, which is called energy. Instances with low energies are assumed to be more likely according to the model. For continuous inverse optimal control, the cost function plays the role of energy function, and the observed expert sequences are assumed to be random samples from the energy-based model so that sequences with low costs are more likely to be observed. We can choose the cost function either as a linear combination of a set of hand-designed features, or a non-linear and non-markovian neural network. The goal is to learn the parameters of the cost function from the expert sequences. 
	
	The parameters can be learned by the maximum likelihood estimation (MLE) in the context of the energy-based model.	The maximum likelihood learning algorithm follows an ``analysis by synthesis'' scheme, which iterates the following two steps: (1) Synthesis step: sample the synthesized trajectories from the current probability distribution using the Markov chain Monte Carlo (MCMC), such as Langevin dynamics \cite{neal2011mcmc}. The gradient computation in the Langevin dynamics can be conveniently and efficiently carried out by back-propagation through time. (2) Analysis step: update the model parameters based on the statistical difference between the synthesized trajectories and the observed trajectories. Such a learning algorithm is very general, and it can learn complex cost functions such as those defined by the neural networks.
	
	We need to point out that MLE is the most commonly used method for learning energy-based models, due to its asymptotic optimality. Among all the asymptotically unbiased estimators, MLE is the most accurate in terms of asymptotic variance \cite{bickel2015mathematical}. Alternative methods for learning EBMs include contrastive divergence (CD) \cite{Hinton02} and noise contrastive estimation (NCE) \cite{hyvarinen2005estimation}. Contrastive divergence replaces MCMC by one or a few steps of MCMC sampling initialized from observed examples, and as a result, it has a big bias.  NCE estimates the energy function discriminatively by recruiting a noise distribution to produce negative or contrastive examples against the observed examples which are treated as positive examples. For accurate estimation, the noise distribution should have substantial overlap with the data distribution. For high dimensional observations such as trajectories, it is difficult to find such a noise distribution. If the noise distribution does not have sufficient overlap with the data distribution, the estimate will have a big variance. Therefore, the maximum likelihood training is a more preferable algorithm to train EBMs.

	For an optimal control problem where the cost function is of the Markovian form, a built-in optimization algorithm is usually already available, such as the iLQR algorithm for autonomous driving. In this case, we also consider a convenient modification of the above learning method, where we change the synthesis step into an optimization step while keeping the analysis step unchanged. We give  justifications for this optimization-based method, although we want to emphasize that the sampling-based method is still more fundamental and principled, and we treat optimization-based method as a convenient modification. 

	Moreover, we propose another novel energy-based IOC framework, where  the energy-based model is trained with a top-down trajectory generator that serves as a fast initializer of the Langevin sampling of the energy-based model through a cooperative learning manner \cite{xie2018cooperative2,xie2018cooperative}. Within each cooperative learning iteration, the trajectory generator generates initial trajectories to initialize a finite-step Langevin dynamics that samples from the energy-based model, and then the energy-based model is trained by comparing the expert trajectories and the synthesized trajectories. After that, the trajectory generator learns from how the MCMC changes its initial generated trajectories. The proposed framework belongs to the ``fast thinking initializer and slow thinking solver'' framework \cite{xie2019cooperative}. The trajectory generator plays the role of the fast thinking initializer because its generation of trajectory is accomplished by direct mapping, while the energy-based model plays the role of the slow thinking solver because it learns a cost function in the form of a conditional energy function, so that the trajectory can be synthesized by minimizing the cost function, or more rigorously by sampling from the energy-based model. The trajectory generator is like a policy, while the energy-based model is like a planner. Compared to GAN-type method, ours is equipped with an iterative refining process (slow thinking) guided by a learned cost (energy)~function.
	
	We empirically demonstrate the proposed energy-based IOC methods on autonomous driving in~both single-agent and multi-agent scenarios, and show that the proposed methods can learn suitable cost functions for optimal control.  

\subsection{Related work}

	The following are research themes related to our work. 
	
	(1) Maximum entropy framework. Our work follows the maximum entropy framework of \cite{ziebart2008maximum} for learning the cost function. Such a framework has also been used previously for generative modeling of images \cite{zhu1998filters} and Markov logic network \cite{richardson2006markov}. In this framework, the energy function is a linear combination of hand-designed features. Recently, \cite{wulfmeier2015maximum} generalized this framework to a deep version. In these methods, the state spaces are discrete, where dynamic programming schemes can be employed to calculate the normalizing constant of the energy-based model. In our work, the state space is continuous, where we use Langevin dynamics via back-propagation through time to sample trajectories from the learned model. We also propose an optimization-based method where we use the gradient descent algorithm or a built-in optimal control algorithm as the inner loop for learning.
	
	(2) ConvNet-EBM. Recently, \cite{xie2019learning_spec, xie2015learning,xie2017synthesizing,xie2020generative,XieZGWZW18,XieXZZW21} applied deep energy-based models to various generative modeling tasks, where the energy functions are parameterized by ConvNets \cite{lecun1995convolutional,krizhevsky2012imagenet}. Our method is different from ConvNet EBM. The control variables in our method form a time sequence. In gradient computation for Langevin sampling, back-propagation through time is used. Also, we propose an optimization-based modification and give justifications. 
	
	(3) Inverse reinforcement learning. Most of the inverse reinforcement learning methods \cite{guided,finn2016connection}, including adversarial learning methods \cite{gan,ho2016generative,li2017infogail,finn2016connection}, involve learning a policy in addition to the cost function. In our work, the energy-based IOC framework (without an extra trajectory generator) does not learn any policy, and it only learns a cost function (i.e., the energy function), where the trajectories are sampled by the Langevin dynamics or obtained by gradient descent or a built-in optimal control algorithm. 
	
	(4) Continuous inverse optimal control (IOC). The IOC problem has been studied by \cite{monfort2015} and \cite{levine2012continuous}. In \cite{monfort2015}, the dynamics is linear and the cost function is quadratic, so that the normalizing constant can be computed by a dynamic programming scheme. In \cite{levine2012continuous}, the Laplace approximation is used for approximation. However, the accuracy of the Laplace approximation is questionable for complex cost function. In our work, we  assume general dynamics and cost function, and we use Langevin sampling for maximum likelihood learning without resorting to Laplace approximation. 
	
	(5) Trajectory prediction. A recent body of research has been devoted to supervised learning for trajectory prediction \cite{socialLSTM,socialGAN,socialattention,desire,deo2018would,Zhao_2019_CVPR}. These methods directly predict the coordinates and do not consider control and dynamic models. Thus, they cannot be used~for~inverse~optimal~control.  
	
	(6) Cooperative Learning. Our joint training framework for IOC follows the generative cooperative learning  algorithm (i.e., the CoopNets algorithm) of \cite{xie2018cooperative} for training the cost function in the EBM and the trajectory generator. Such a learning algorithm has also been applied previously to image generation \cite{xie2018cooperative}, video generation \cite{xie2018cooperative}, 3D shape generation \cite{XieZGWZW18}, supervised conditional learning \cite{xie2019cooperative}, and unsupervised image-to-image translation \cite{XieZFZW21}. The CoopVAEBM \cite{XieZL21} is a variant of the CoopNets algorithm by replacing the generic generator with a variational auto-encoder (VAE) \cite{KingmaW13}. The CoopFlow \cite{xie2022coopflow} is another variant of the CoopNets algorithm by changing the generator into a normalizing flow \cite{KingmaD18}.   

\subsection{Contributions}

	The contributions of our work are as follows. 
	
\begin{itemize}

	\item We propose an energy-based method for continuous inverse optimal control based on Langevin sampling via back-propagation through time. To the best of our knowledge, this is the first work that studies MCMC sampling-based inverse optimal control. Such an ``analysis by synthesis'' learning scheme makes our work essentially different from \cite{ziebart2008maximum,monfort2015,levine2012continuous}. 
	\item We also propose an optimization-based method as a convenient approximation of the MCMC sampling under the proposed energy-based learning framework. The modified algorithm becomes an ``analysis by optimization'' scheme. 
	\item We evaluate the proposed methods on autonomous driving tasks for trajectory prediction. We apply our framework to both single-agent system and multi-agent system, with both linear cost function and neural network non-linear cost function. This is the first work to study vehicle trajectory prediction under the energy-based framework.  
	\item We also propose to train an energy-based model together with a policy-like trajectory generator, which serves as a fast initializer for the Langevin sampling, in a cooperative learning scheme. 
	\item We conduct extensive ablation studies to analyze the effects of the key components and hyperparameters of the proposed frameworks to understated the model behaviors.  
	
\end{itemize}
	
\subsection{Organization}
The rest of our paper is organized as follows: Section \ref{sec:EBIPC} presents the proposed framework of the energy-based inverse optimal control. Section \ref{sec:coop} presents the proposed joint training framework, in which the energy-based model is trained simultaneously with a trajectory generator as amortized sampler. Qualitative and quantitative results of experiments  are shown in Section \ref{sec:experients}. Conclusion of the paper is given in Section \ref{sec:con}.

\section{Energy-based inverse optimal control}\label{sec:EBIPC}

\subsection{Optimal control} 

We study the finite horizon control problem for discrete time $t \in \{1, ..., T\}$. Let $x_t$ be the state at time $t$. Let ${\bf x} = (x_t, t = 1, ..., T)$. Let $u_t$ be the continuous control variable or action at time $t$. Let ${\bf u} = (u_t, t = 1, ..., T)$. The dynamics is assumed to be deterministic, $x_{t} = f(x_{t-1}, u_t)$, where $f$ is given, so that ${\bf u}$ determines ${\bf x}$. The trajectory is $({\bf x}, {\bf u}) = (x_t, u_t, t = 1, ..., T)$. Let $e$ be the environment condition. We assume that the recent history $h = (x_t, u_t, t = -k, ..., 0)$ is known. 

The cost function is $C_\theta({\bf x}, {\bf u}, e, h)$ where $\theta$ consists of the parameters that define the cost function. Its special case is of the linear form $C_\theta({\bf x}, {\bf u}, e, h) = \langle \theta, \phi({\bf x}, {\bf u}, e, h)\rangle$,	where $\phi$ is a vector of hand-designed features, and $\theta$ is a vector of weights for these features. We can also parameterize $C_\theta$ by a neural network. The problem of optimal control is to find ${\bf u}$ to minimize $C_\theta({\bf x}, {\bf u}, e, h)$ with given $e$ and $h$ under the known dynamics $f$. The problem of inverse optimal control is to learn $\theta$ from expert demonstrations $D = \{({\bf x}_i, {\bf u}_i, e_i, h_i), i = 1, ..., n\}$. 

\subsection{Energy-based probabilistic model}

The energy-based model assumes the following conditional probability density function 
\begin{align} 
 p_\theta({\bf u}|e, h) = \frac{1}{Z_\theta(e, h)} \exp[ - C_\theta({\bf x}, {\bf u}, e, h) ], \label{eq:e}
\end{align}
where $Z_\theta(e, h) = \int \exp[ - C_\theta({\bf x}, {\bf u}, e, h) ] d {\bf u}$ is the normalizing constant. Recall that ${\bf x}$ is determined by ${\bf u}$ according to the deterministic dynamics, so that we only need to define probability density on ${\bf u}$. The cost function $C_\theta$ serves as the energy function. For expert demonstrations $D$, ${\bf u}_i$ are assumed to be random samples from $p_\theta({\bf u} | e_i, h_i)$, so that ${\bf u}_i$ tends to have low cost $C_\theta({\bf x}, {\bf u}, e_i, h_i)$. 

\subsection{Sampling-based inverse optimal control} 

The parameters $\theta$ can be learned by maximum likelihood. The log-likelihood is given by
\begin{eqnarray}
  L(\theta) = \frac{1}{n} \sum_{i=1}^{n} \log p_\theta({\bf u}_i | e_i, h_i). 
\end{eqnarray} 
We can maximize $L(\theta)$ by gradient ascent, and the learning gradient is computed by
\begin{eqnarray}
\begin{aligned}
  L'(\theta) = \frac{1}{n} \sum_{i=1}^{n}\bigg[&\E_{p_\theta({\bf u}| e_i, h_i)}\left( \frac{\partial}{\partial \theta} C_\theta({\bf x}, {\bf u}, e_i, h_i) \right) \\ 
 &- \frac{\partial}{\partial \theta} C_\theta({\bf x}_i, {\bf u}_i, e_i, h_i) \bigg], \label{eq:l1} 
 \end{aligned}
 \end{eqnarray}
 which follows the property of the normalizing constant $\frac{\partial}{\partial \theta} \log Z_\theta(e, h) =- \E_{p_\theta({\bf u}| e, h)}\left( \frac{\partial}{\partial \theta} C_\theta({\bf x}, {\bf u}, e, h) \right)$.
 
In order to approximate the above expectation, we can generate multiple random samples via $\tilde{\bf u}_i \sim p_\theta({\bf u} | e, h)$, which generates each sampled trajectory $(\tilde{\bf x}_i, \tilde{\bf u}_i)$ by unfolding the dynamics. We estimate $L'(\theta)$ by 
\begin{align} 
	\hat{L}'(\theta) = \frac{1}{n} \sum_{i=1}^{n}\bigg[ & \frac{\partial}{\partial \theta} C_\theta(\tilde{\bf x}_i, \tilde{\bf u}_i, e_i, h_i) -  \frac{\partial}{\partial \theta} C_\theta({\bf x}_i, {\bf u}_i, e_i, h_i) \bigg], \label{eq:lg}
 \end{align}
 which is the stochastic unbiased estimator of $L'(\theta)$. Then we can run the gradient ascent algorithm $\theta_{\tau+1} = \theta_\tau + \gamma_\tau \hat{L}'(\theta_\tau)$ to obtain the maximum likelihood estimate of $\theta$, where $\tau$ indexes the time step, $\gamma_\tau$ is the step size. According to the Robbins-Monroe theory of stochastic approximation ~\cite{robbins1951stochastic}, if $\sum_\tau \gamma_\tau = \infty$ and $\sum_\tau \gamma_\tau^2 < \infty$, the algorithm will converge to a solution of $L'(\theta) = 0$. For each $i$, we can also generate multiple copies of $(\tilde{\bf x}_i, \tilde{\bf u}_i)$ from $p_\theta({\bf u}| e_i, h_i)$ and average them to approximate the expectation in Equation~(\ref{eq:l1}). A small number is sufficient because the averaging effect takes place over time. 

 In linear case, where 
$ 
 C_\theta({\bf x}, {\bf u}, e, h) = \langle \theta, \phi({\bf x}, {\bf u}, e, h)\rangle,
$
 we have $\frac{\partial}{\partial \theta} C_\theta({\bf x}, {\bf u}, e, h) = \phi({\bf x}, {\bf u}, e, h)$, making 
 $\hat{L}'(\theta) = \frac{1}{n} \sum_{i=1}^{n}\left[\phi(\tilde{\bf x}_i, \tilde{\bf u}_i, e_i, h_i) - \phi({\bf x}_i, {\bf u}_i, e_i, h_i)\right]$. 
 It is the statistical difference between the observed trajectories and synthesized trajectories. At maximum likelihood estimate, the two match each other. 
 
The synthesis step that samples from $p_\theta({\bf u} |e, h)$ can be accomplished by an efficient gradient-based MCMC, the Langevin dynamics, which iterates the following steps: 
\begin{align} 
	{\bf u}_{s+1} &= {\bf u}_s + \frac{\delta^2}{2} \frac{\partial}{\partial {\bf u}} \log p_{\theta}({\bf u}_s|e,h) + \delta {\bf z}_s, \nonumber\\
&= {\bf u}_s - \frac{\delta^2}{2} \frac{\partial}{\partial {\bf u}} C_\theta({\bf x}_s, {\bf u}_s, e, h) + \delta {\bf z}_s, \label{eq:lang}
\end{align}
where $s$ indexes the time step, $\delta$ is the step size, and ${\bf z}_s\sim {\mathcal N}(0, I)$ the Brownian motion independently over $s$, where $I$ is the identity matrix of the same dimension as ${\bf u}$. The Langevin dynamics is an inner loop of the learning algorithm, with ${\bf u}_{0}$ (${\bf u}$ at the initial time step) being initialized by Gaussian white noise.  The gradient descent part ${\bf u}_{s+1} = {\bf u}_s - \frac{\delta^2}{2} \frac{\partial}{\partial {\bf u}} C_\theta({\bf x}_s, {\bf u}_s, e, h)$ of Equation~(\ref{eq:lang}) is a mode seeking process that minimizes the cost function $C_{\theta}$, while the added Gaussian noise ${\bf z}_s$ will prevent the samples from being trapped by local minima. According to the second law of thermodynamics \cite{cover2006elements}, as $s \rightarrow \infty$ and $\delta \rightarrow 0$, ${\bf u}_{s}$ becomes an exact sample from $p_{\theta}({\bf u}|e,h)$ under some regularity conditions. A Metropolis-Hastings \cite{hastings1970monte} step can also be added to correct for the error due to discrete time step in Equation (\ref{eq:lang}), but most  existing works, such as \cite{ChenFG14,xie2016theory,NijkampH0ZW20}, have shown that this can be ignored in practice if a small enough step size $\delta$ is used. Thus, for computational efficiency, in this work, we do not have the Metropolis-Hastings correction in our implementation.

The gradient term $\partial C_\theta({\bf x}, {\bf u}, e, h)/\partial {\bf u}$ is computed via back-propagation through time, where ${\bf x}$ can be obtained from ${\bf u}$ by unrolling the deterministic dynamics. The computation can be efficiently and conveniently carried out by auto-differentiation on the current deep learning platforms. 

\subsection{Optimization-based inverse optimal control} 

We can remove the noise term in Langevin dynamics in Equation (\ref{eq:lang}), to make it a gradient descent process, i.e., ${\bf u}_{s+1} = {\bf u}_s - \eta \frac{\partial}{\partial {\bf u}} C_\theta({\bf x}_s, {\bf u}_s, e, h)$, and we can still learn the cost function that enables optimal control. This amounts to modifying the synthesis step into an optimization step. 
	
Moreover, a built-in optimization algorithm is usually already available for minimizing the cost function $C_\theta({\bf x}, {\bf u}, e, h)$ over ${\bf u}$. For instance, in autonomous driving, a commonly used algorithm is iLQR. In this case, we can replace the synthesis step by an optimization step, where, instead of sampling $\tilde{\bf u}_i \sim p_{\theta_t}({\bf u} | e_i, h_i)$, we optimize 
\begin{align}
\tilde{\bf u}_i = \arg\min_{\bf u} C_\theta({\bf x}, {\bf u}, e_i, h_i). \label{eq:o}
\end{align}
The analysis step remains unchanged. In this paper, we emphasize the sampling-based method, which is more principled maximum likelihood learning, and we treat the optimization-based method as a convenient modification. We will evaluate both learning methods in our experiments.

A justification for the optimization-based algorithm in the context of the energy-based model in Equation~(\ref{eq:e}) is to consider its tempered version $p_\theta({\bf u}| e, h) \propto \exp[-C_\theta({\bf x}, {\bf u}, e, h)/T]$, where $T$ is the temperature. Then the optimized $\tilde{\bf u}$ that minimizes $C_\theta({\bf x}, {\bf u}, e, h)$ can be considered the zero-temperature sample, which is used to approximate the expectation in Equation~(\ref{eq:l1}).

\textbf{Moment matching}. For simplicity, consider the linear cost function $C_\theta({\bf x}, {\bf u}, e, h) = \langle \theta, \phi({\bf x}, {\bf u}, e, h)\rangle$. At the convergence of the optimization-based learning algorithm, which has the same analysis step as the sampling-based algorithm, we have $\hat{L}'(\theta) = 0$, so that 
 \begin{align}
 \frac{1}{n} \sum_{i=1}^{n}\phi(\tilde{\bf x}_i, \tilde{\bf u}_i, e_i, h_i) = \frac{1}{n} \sum_{i=1}^{n}\ \phi({\bf x}_i, {\bf u}_i, e_i, h_i),  
 \end{align} 
where the left-hand side is the average of the optimal behaviors obtained by Equation~(\ref{eq:o}), and the right-hand side is the average of the observed behaviors. 
We want the optimal behaviors to match the observed behaviors on average. 
We can see the above point most clearly in the extreme case where all $e_i = e$ and all $h_i = h$, so that $\phi(\tilde{\bf x}, \tilde{\bf u}, e, h) = \frac{1}{n} \sum_{i=1}^{n}\ \phi({\bf x}_i, {\bf u}_i, e, h)$, i.e., we want the optimal behavior under the learned cost function to match the average observed behaviors as far as the features of the cost function are concerned. Note that the matching is not in terms of raw trajectories but in terms of the features of the cost function. In this matching, we do not care about modeling the variabilities in the observed behaviors. In the case of different $(e_i, h_i)$ for $i = 1, ..., n$, the matching may not be exact for each combination of $(e, h)$. However, such mismatches may be detected by new features which can be included in the features of the cost function.

\textbf{Adversarial learning}. We can also justify this optimization-based algorithm outside the context of probabilistic model as adversarial learning. To this end, we re-think about the inverse optimal control, whose goal is not to find a probabilistic model for the expert trajectories. Instead, the goal is to find a suitable cost function for optimal control, where we care about the optimal behavior, not the variabilities of the observed behaviors. Define the value function 
\begin{align}
 V(\theta,\{\tilde{\bf u}_i\}) = \frac{1}{n} \sum_{i=1}^{n}\left[C_\theta(\tilde{\bf x}_i, \tilde{\bf u}_i, e_i, h_i) - C_\theta({\bf x}_i, {\bf u}_i, e_i, h_i)\right], 
\end{align} 
then $\hat{L}'(\theta) = \frac{\partial}{\partial \theta} V$, so that the analysis step increases $V$. The optimization step and the analysis step play an adversarial game $\max_\theta \min_{\tilde{\bf u}_i, \forall i}V$, where the optimization step seeks to minimize $V$ by reducing the costs, while the analysis step seeks to increase $V$ by modifying the cost function. More specifically, the optimization step finds the minima of the cost functions to decrease $V$, whereas the analysis step shifts the minima toward the observe trajectories in order to increase $V$. 

\subsection{Energy-based IOC algorithm}

Algorithm \ref{alg:1} and Algorithm \ref{alg:2} present the sampling-based and optimization-based learning algorithms, respectively. We treat the sampling-based method as a more fundamental and principled method, and the optimization-based method as a convenient modification.  In our experiments, we shall evaluate both sampling-based method using Langevin dynamics and optimization-based method with gradient descent (GD) or iLQR as optimizer. 

\begin{algorithm} 
	\caption{Energy-based IOC with synthesis step}\label{alg:1}
	\begin{algorithmic}[1]
		\State \textbf{input} expert demonstrations {$D = \{({\bf x}_i, {\bf u}_i, e_i, h_i), \forall i\}$}.
		\State \textbf{output} cost function parameters $\theta$, and synthesized trajectories $\{(\tilde{\bf x}_i, \tilde{\bf u}_i), \forall i\}$.
		\State Let $\tau$ $\gets$ 0, randomly initialize $\theta$.
		\Repeat
		\State {\bf synthesis step}: for each $i$, synthesize $\tilde{\bf u}_i \sim p_{\theta_t}({\bf u} | e_i, h_i)$ by Langevin sampling and then obtain $\tilde{\bf x}_i$ . 
		\State {\bf analysis step}: update $\theta_{\tau+1} = \theta_\tau + \gamma_\tau \hat{L}'(\theta_\tau)$, where $\hat{L}'$ is computed according to Equation~(\ref{eq:lg}).
		\State $\tau$ $\gets$ $\tau + 1$.
		\Until{$\tau = \tau_{\rm max}$, the number of iterations.}
	\end{algorithmic}
\end{algorithm}

	\begin{algorithm} 
	\caption{Energy-based IOC with optimization step}\label{alg:2}
	\begin{algorithmic}[1]
		\State \textbf{input} expert demonstrations {$D = \{({\bf x}_i, {\bf u}_i, e_i, h_i), \forall i\}$}.
		\State \textbf{output} cost function parameters $\theta$, and optimized trajectories $\{(\tilde{\bf x}_i, \tilde{\bf u}_i), \forall i\}$.
		\State Let $\tau$ $\gets$ 0, randomly initialize $\theta$.
		\Repeat
		\State {\bf optimization step}:  for each $i$, optimize $\tilde{\bf u}_i = \arg\min_{\bf u} C_\theta({\bf x}, {\bf u}, e_i, h_i)$, by gradient descent (GD) or iLQR, and then obtain $\tilde{\bf x}_i$. 
		\State {\bf analysis step}: update $\theta_{\tau+1} = \theta_\tau + \gamma_\tau \hat{L}'(\theta_\tau)$, where $\hat{L}'$ is computed according to Equation~(\ref{eq:lg}).
		\State $\tau$ $\gets$ $\tau + 1$.
		\Until{$\tau = \tau_{\rm max}$, the number of iterations.}
	\end{algorithmic}
\end{algorithm}
	
\section{Joint training}\label{sec:coop}

\subsection{A trajectory generator model as a fast initializer}
	
Both sampling-based method via Langevin dynamics and optimization-based method via gradient descent are based on iterative process, which will benefit from good initialization. A good initial point can not only greatly shorten the number of iterative steps but also help find the optimal modes of the cost function. Therefore, we propose to train an energy-based model simultaneously with a trajectory generator model that serves as a fast initializer for the Langevin dynamics or gradient descent of the energy-based model. 

The basic idea is to use the trajectory generator model to generate trajectories via ancestral sampling to initialize a finite step Langevin dynamics or gradient descent for training the energy-based model. In return, the trajectory generator model learns from how the Langevin dynamics or gradient descent updates the initial trajectories it generates. Such a cooperative learning strategy is proposed in \cite{xie2018cooperative,xie2019cooperative,xie2022coopflow} for image generation. 

To be specific, we propose the trajectory generator model that consists of the following two components
\begin{align}
u_{t} &= F_{\alpha}(x_{t-1}, \xi_t, e) \label{eq:policy} \\
x_{t} &= f(x_{t-1}, u_{t}) \label{eq:dynamic}
\end{align}
where $t = 1, ..., T$, Equation~(\ref{eq:policy}) is the policy model, and Equation~(\ref{eq:dynamic}) is the
known dynamic function. $\xi_t \sim \mathcal{N}(0,I)$ is the Gaussian noise vector. The Gaussian noise vectors at different times $(\xi_t, t = 1, ..., T)$ are independent of each other. Given the state $x_{t-1}$ at the previous time step $t-1$ along with the environment condition $e$, the policy model outputs the action $u_{t}$ at the current time step $t$, where the noise vector $\xi_t$ accounts for the randomness in the mapping from $x_{t-1}$ to $u_{t}$. $F_{\alpha}$ is a multi-layer perceptron, where $\alpha$ is the  model parameters of the network. The initial state $x_0$ is assumed to be given.  

We denote ${\boldsymbol \xi}=(\xi_t, t=1,...,T)$ and $p(\boldsymbol \xi) = \prod_{t=1}^{T} p(\xi_t)$. Given the state $x_{t-1}$ and the environment condition $e$, although $x_t$ is dependent on the action $u_t$, $u_t$ is generated from $\xi_t$. In fact, we can write the trajectory generator in a compact form, i.e., ${\bf u} = G_{\alpha}({\boldsymbol \xi}, e, h)$, where $G_{\alpha}$ composes $F_{\alpha}$ and $f$ over time, and we use $h=x_0$ for simplicity in our implementation. 
	
The algorithm for joint training of the energy-based model and the trajectory generator is that: at each iteration, (i) we first sample ${\boldsymbol \xi}_i$ from
the Gaussian prior distribution, and then generate the initial trajectories by $\hat{{\bf u}}_i= G_{\alpha}({\boldsymbol \xi}_i, e_i, h_i)$ for $i=1,...,n$. (ii) Starting from the initial trajectories $\{\hat{{\bf u}}_i\}$, we sample from the energy-based model by running a finite number of Langevin steps or optimize the cost function by running a finite steps of gradient descent to obtain the updated trajectories $\{\tilde{{\bf u}}_i\}$, and then obtain $\{\tilde{{\bf x}}_i\}$. (iii) We update the parameters $\theta$ of the energy-based model by maximum likelihood estimation, where the computation of the gradient of the likelihood is based on $\{\tilde{{\bf u}}_i\}$ and follows Equation (\ref{eq:lg}). (iv) We update the parameters $\alpha$ of the trajectory generator by gradient descent on the loss
\begin{align}
\hat{l}_{g}'(\alpha) =   \frac{\partial}{\partial \alpha} \left[ \frac{1}{n}\sum_{i=1}^{n}||\tilde{{\bf u}}_i - G_{\alpha}({\boldsymbol \xi}_i, e_i, h_i)||^2\right]. 
\label{eq:loss_G}
\end{align}
Algorithm \ref{alg:3} presents a detailed description of the cooperative training algorithm of an energy-based model and a trajectory generator for inverse optimal control. Synthesis step and optimization step are two options to generate $(\tilde{{\bf u}},\tilde{{\bf x}})$. 
	
	\begin{algorithm} 
		\caption{Energy-based IOC with a trajectory generator}\label{alg:3}
		\begin{algorithmic}[1]
			\State \textbf{input} expert demonstrations {$D = \{({\bf x}_i, {\bf u}_i, e_i, h_i), \forall i\}$}.
			\State \textbf{output} cost function parameters $\theta$, trajectory generator parameters $\alpha$, and synthesized or optimized trajectories $\{(\tilde{\bf x}_i, \tilde{\bf u}_i), \forall i\}$.
			\State Let $\tau$ $\gets$ 0, randomly initialize $\theta$ and $\alpha$.
			\Repeat
             \State {\bf Initialization step}: Initialize $\hat{\bf u}_i = G_{\alpha_t}(\boldsymbol \xi_i, e_i, h_i)$, where $\boldsymbol \xi_i \sim p({\boldsymbol \xi})$ by ancestral sampling, and then obtain $\hat{\bf x}_i$ for each $i$.  				
			\State {\bf Synthesis step or optimization step}: Given the initial $\hat{\bf u}_i$, synthesizing $\tilde{\bf u}_i \sim p_{\theta_t}({\bf u} | e_i, h_i)$ by Langevin sampling, or optimizing $\tilde{\bf u}_i = \arg\min_{\bf u} C_\theta({\bf x}, {\bf u}, e_i, h_i)$ by gradient descent (GD) or iLQR, and then obtain $\tilde{\bf x}_i$, for each $i$. 
			\State {\bf Analysis step (update cost function)}: Update $\theta_{\tau+1} = \theta_\tau + \gamma_\tau \hat{L}'(\theta_\tau)$, where $\hat{L}'(\theta)$ is computed according to Equation~(\ref{eq:lg}).
			\State {\bf Analysis step (update policy model)}: Update $\alpha_{\tau+1} = \alpha_\tau - \eta_\tau \hat{l}_{g}'(\alpha_\tau)$, where $\hat{l}_{g}'(\alpha)$ is computed according to Equation~(\ref{eq:loss_G}).
			\State $\tau$ $\gets$ $\tau + 1$.
			\Until{$\tau = \tau_{\rm max}$, the number of iterations.}
		\end{algorithmic}
	\end{algorithm}	
	
\subsection{Bottom-up and top-down generative models of trajectories}
Algorithm \ref{alg:3} presented in the main text is about a joint training of two types of generative models, the energy-based model (we also call it the trajectory evaluator) and the latent variable model (i.e., the trajectory generator). Both of these two models can be parameterized by deep neural networks and they are of opposite directions. The energy-based model has a bottom-up energy function that maps the trajectory to the cost, while the trajectory generator owns a top-down transformation that maps the sequence of noise vectors (i.e., the latent variables) to the trajectory, as illustrated by the following diagram.

\begin{eqnarray}
\begin{array}[c]{ccc}
\mbox{{ Bottom-up model}} && \mbox{{ Top-down model}}\\
\mbox{{\bf cost}} & & \mbox{{\bf  noise sequence}}\\
\mbox{{\bf (energy)}} & & \mbox{{\bf  (latent variables)}}\\
\Uparrow&&\Downarrow\\
{\rm trajectory} & & {\rm trajectory}\\
\mbox{(a) Trajectory evaluator}  &&\mbox{ (b) Trajectory generator}\\
\mbox{(energy-based model)}  &&\mbox{(latent variable model)}
\end{array} \nonumber  \label{eq:diagram0}
\end{eqnarray}

\subsection{Iterative and non-iterative generations of trajectories}
The energy-based model $p_\theta({\bf u}|e, h)$ defines a cost function or an energy function $C_\theta({\bf x}, {\bf u}, e, h)$, from which we can derive the Langevin dynamics to generate ${\bf u}$. This is an implicit generation process of ${\bf u}$ that iterates the Langevin step in Equation~(\ref{eq:lang}).  

\begin{figure}[h]
	\centering
	\includegraphics[height=0.12\textwidth]{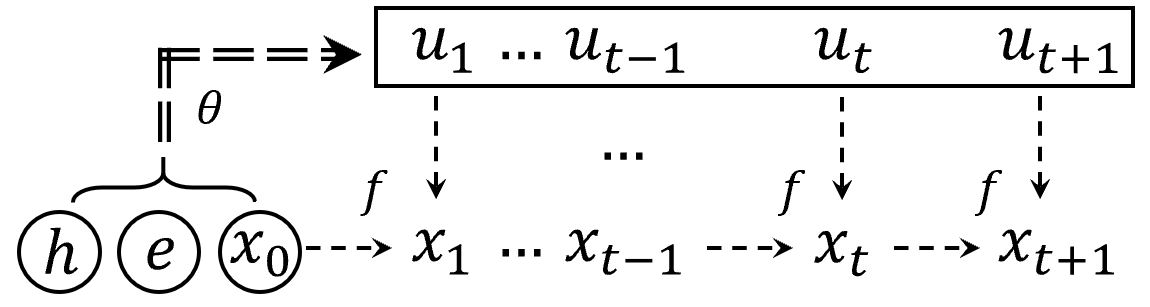} \hspace{7mm}
	(a) Trajectory generation via Langevin sampling \hspace{7mm} 
	\includegraphics[height=0.12\textwidth]{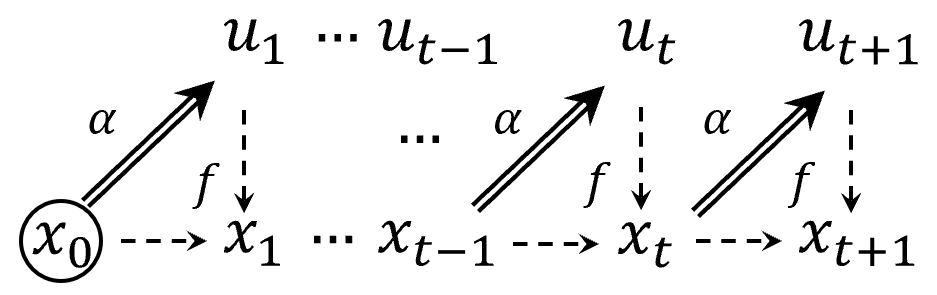}\\
	(b) Trajectory generation via ancestral sampling
	\caption{Trajectory generation by (a) iterative method and (b) non-iterative method.}\label{fig:generation}
\end{figure}

Figure \ref{fig:generation} (a) illustrates the generation process of $({\bf u},{\bf x})$. Given $(h,e)$, the Langevin sampling seeks to find ${\bf u}=(u_1,...,u_T)$ to minimize the $C_\theta({\bf x}, {\bf u}, e, h)$. The dashed~double line arrows indicate iterative generation by sampling in the energy-based model, while the dashed
line arrows indicate the known dynamic function $x_t = f(x_{t-1}, u_t)$. With the generated action sequence ${\bf u}=(u_1,...,u_T)$, the state sequence ${\bf x}=(x_1,...,x_T)$ can be easily obtained by applying~the~dynamic~function.

The trajectory generator generates ${(\bf u},{\bf x})$ via ancestral sampling, $({\bf u},{\bf x})=G_{\alpha}(\boldsymbol \xi, e,h)$ which is a non-iterative process to produce ${(\bf u},{\bf x})$ from the recent history $h$ (We assume $h = x_0$), environment $e$, and a sequence of noise vectors  ${\boldsymbol \xi}=(\xi_1,...\xi_T)$ serving as the latent variables. The generator can unfold over time and can be decomposed into the policy model $u_t=F_{\alpha}(x_{t-1},\xi_t,e)$ and the dynamic function $x_t = f(x_{t-1}, u_t)$ at each time step. The latent variable $\xi_t$ accounts for variation in the policy model at time step $t$. Figure \ref{fig:generation}(b) illustrates the generation process of the trajectory generator. The double line arrows indicate the mapping of the policy model, while the dashed line arrows indicate the known dynamic function. The whole process of generating $({\bf u},{\bf x})$ is
of a dynamic or causal nature in that it directly evolves or unfolds over time.

\subsection{Issue in maximum likelihood training of a single trajectory generator}
Let ${\boldsymbol \xi}=(\xi_t, t=1,...,T)$, where $\xi_t \sim \mathcal{N}(0,I)$ independently over time $t$. Let ${\bf u}=(u_t,t=1,...,T)$. We have ${\bf u}=G_{\alpha}({\boldsymbol \xi},e,h)+{\boldsymbol \epsilon}$, where ${\boldsymbol \epsilon}=(\epsilon_t,t=1,...,T)$ are observation errors and $\epsilon_t \sim \mathcal{N}(0,\sigma^2I)$. For notational simplicity, we omit ${\bf x}$ and only keep ${\bf u}$ in the output of $G_{\alpha}$, because ${\bf x}$ is just the intermediate output of $G_{\alpha}$ and ${\bf u}$ is determined by ${\bf x}$. The trajectory generator defines the joint distribution of $({\bf u}, {\boldsymbol \xi})$ conditioned on $(e,h)$ as below,
\begin{equation}
q_\alpha({\bf u}, {\boldsymbol \xi}|e,h) = q_{\alpha}({\bf u}| {\boldsymbol \xi},e,h) p({\boldsymbol \xi})
\end{equation}
where $p({\boldsymbol \xi})=\prod_{i=t}^{T}p(\xi_t)$ is the prior distribution and $q_{\alpha}({\bf u}|{\boldsymbol \xi},e,h)=\mathcal{N}(G_{\alpha}({\boldsymbol \xi},e,h),\sigma^2 I)$. The marginal distribution of ${\bf u}$ conditioned on $(e,h)$ is given by $q_{\alpha}({\bf u}|e,h)=\int q_{\alpha}({\bf u},{\boldsymbol \xi} |e,h)d{\boldsymbol \xi}$. The posterior distribution is $q_{\alpha}({\boldsymbol \xi}|{\bf u}, e, h)=q_{\alpha}({\bf u}, {\boldsymbol \xi}|e,h)/q_{\alpha}({\bf u}|e,h)$. Suppose we observe expert demonstrations $D=\{({\bf u}_i,{\bf x}_i,e_i,h_i),i=1,...,n\}$. The maximum likelihood estimation of $\alpha$ seeks to maximize the log-likelihood function
\begin{equation}
L(\alpha)=\sum_{i=1}^n \log q_{\alpha}({\bf u_i}|e_i,h_i). \label{eqn:likelihood_G}
\end{equation}
The learning gradient can be computed  according to
\begin{align}
&\frac{\partial}{\partial \alpha} \log q_{\alpha}({\bf u}|e,h)= \frac{1}{q_{\alpha}({\bf u}|e,h)} \frac{\partial}{\partial \alpha} \int q_{\alpha}({\bf u}, {\boldsymbol \xi}|e,h) d {\boldsymbol \xi}\\
=&\int \left[ \frac{\partial}{\partial \alpha} \log q_{\alpha}({\bf u},{\boldsymbol \xi}|e,h)\right] \frac{q_{\alpha}({\bf u}, {\boldsymbol \xi}|e,h)}{q_{\alpha}({\bf u} |e,h)} d {\boldsymbol \xi} \\
=&\E_{q_{\alpha}({\boldsymbol \xi}|{\bf u},e,h)}\left[\frac{\partial}{\partial \alpha} \log q_{\alpha}({\bf u},{\boldsymbol \xi}|e,h)\right] \label{eqn:gradient}
\end{align}
The expectation term in Equation (\ref{eqn:gradient}) is under the posterior distribution $q_{\alpha}({\boldsymbol \xi}|{\bf u},e,h)$ that is analytically intractable. One may draw samples from $q_{\alpha}({\boldsymbol \xi}|{\bf u},e,h)$ via Langevin inference dynamics that iterates 
\begin{align}
{\boldsymbol \xi}_{s+1} = {\boldsymbol \xi}_{s} + \frac{\delta^2}{2} \frac{\partial}{\partial {\boldsymbol \xi}} \log q_{\alpha}({\boldsymbol \xi}_s|{\bf u},e,h) + \delta {\bf z}_s,\label{eqn:inference_G}
\end{align}
where ${\bf z} \sim \mathcal{N}(0,I)$, $\delta$ is the step size and $s$ indexes the time step. Also, $\xi_0$ is usually sampled from Gaussian white noise for initialization. After we infer ${\boldsymbol \xi}$ from each observation $({\bf u}_i,e_i,h_i)$ by sampling from $q_{\alpha}({\boldsymbol \xi}|{\bf u}_i,e_i,h_i)$ via Langevin inference process, the Monte Carlo approximation
of the gradient of $L(\alpha)$ in Equation (\ref{eqn:likelihood_G}) is computed by
\begin{align}
\frac{\partial}{\partial \alpha}L(\alpha) \approx \sum_{i=1}^{n} \left[\frac{\partial}{\partial \alpha} \log q_{\alpha}({\bf u}_i,{\boldsymbol \xi}_i|e_i,h_i)\right] \label{eqn:learning_G}
\end{align}
Since 
$\frac{\partial}{\partial {\boldsymbol \xi}} \log q_{\alpha}({\boldsymbol \xi}|{\bf u},e,h)=\frac{\partial}{\partial {\boldsymbol \xi}} \log q_{\alpha}({\bf u},{\boldsymbol \xi}|e,h)$ in Equation (\ref{eqn:inference_G}). Both inference step in Equation (\ref{eqn:inference_G}) and learning step in Equation (\ref{eqn:learning_G}) need to compute derivative of  $\log q_{\alpha}({\bf u},{\boldsymbol \xi}|e,h)=\frac{1}{2\sigma^2}||G_{\alpha}({\boldsymbol \xi},e,h)-{\bf u}||^2+ {\textrm{const}} $. The former is with respect to ${\boldsymbol \xi}$, while the latter is with respect to $\alpha$,  both of which can be computed by back-propagation through time. The resulting algorithm is called alternating back-propagation through time (ABPTT) algorithm \cite{xie2019learningrep}.

Although the ABPTT algorithm is natural and simple, the difficulty of training the trajectory generator in this way might lie in the non-convergence issue of the short-run Langevin inference in Equation (\ref{eqn:inference_G}). Even long-run Langevin inference chains are easy to get trapped by local modes. Without fair samples drawn from the posterior distribution, the estimation of $\alpha$ will be biased.   

\subsection{Understanding the learning behavior of the cooperative training}
In this section, we will present a theoretical understand of the learning behavior of the proposed Algorithm 2 shown in main text. We firstly start from the Contrastive Divergence (CD) algorithm that was proposed for efficient training of  energy-based models. The CD runs $k$ steps of MCMC initialized from the training examples, instead of the Gaussian white noise. Given the energy-based model for IOC $p_{\theta}({\bf u}|e,h)$. Let $M_{\theta}$ be the transition kernel of the finite-step MCMC that samples from $p_{\theta}({\bf u}|e,h)$. The original CD learning of $p_{\theta}({\bf u}|e,h)$ seeks to minimize 
\begin{eqnarray}
{\theta}_{\tau+1} = \arg \min_{\theta} [ \KL(p_{\rm expert}({\bf u}|e,h){\parallel}p_{\theta}({\bf u}|e,h)) - \nonumber \\ 
\KL(M_{{\theta}_{\tau}} p_{\rm expert}({\bf u}|e,h){\parallel} p_{\theta}({\bf u}|e,h))],\label{eqn:CD}
\end{eqnarray}
where $p_{\rm expert}({\bf u}|e,h)$ is the unknown distribution of the observed demonstrations of experts. Let $M_{\theta} p_{\rm expert}({\bf u}|e,h)$ denote the 
marginal distribution obtained after running $M_{\theta}$ starting from $p_{\rm expert}({\bf u}|e,h)$. If $M_{\theta} p_{\rm expert}({\bf u}|e,h)$ converges to~$p_{\theta}({\bf u}|e,h)$, then the second KL-divergence will become very small, and~the CD estimate eventually is close to maximum likelihood estimate which minimizes the first KL-divergence in Equation~(\ref{eqn:CD}).

In Algorithm \ref{alg:3}, the MCMC sampling of the energy-based model is initialized from the trajectory generator $q_{\alpha}({\bf u}|e,h)$, thus the learning of the energy-based model follows a modified CD estimate which, at learning step $\tau$, seeks to minimize
\begin{eqnarray}
{\theta}_{\tau+1} = \arg \min_{\theta} [ \KL(p_{\rm expert}({\bf u}|e,h){\parallel}p_{\theta}({\bf u}|e,h)) - \nonumber  \\  
\KL(M_{{\theta}_{\tau}} q_{\alpha}({\bf u}|e,h){\parallel} p_{\theta}({\bf u}|e,h))],\label{eqn:CD2}
\end{eqnarray}
where we replace the $p_{\rm expert}({\bf u}|e,h)$ in Equation (\ref{eqn:CD2}) by $q_{\alpha}({\bf u}|e,h)$. That means we run a finite-step MCMC from a given initial distribution $q_{\alpha}({\bf u}|e,h)$, and use the resulting samples as synthesized examples to approximate the gradient of the log-likelihood of the EBM.

At learning step $\tau$, the learning of $q_{\alpha}({\bf u}|e,h)$ seeks to minimize
\begin{eqnarray}
{\alpha}_{\tau+1} = \arg \min_{\alpha} [  \KL(M_{{\theta}} q_{{\alpha}_{\tau}}({\bf u}|e,h){\parallel} q_{\alpha}({\bf u}|e,h))].\label{eqn:CD_G}
\end{eqnarray}
Equation (\ref{eqn:CD_G}) shows that $q_{\alpha}$ learns to be the stationary distribution of $M_{\theta}$. In other words, $q_{\alpha}$ seeks to be close to $p_{\alpha}$, i.e., $q_{\alpha} \rightarrow p_{\theta}$. If so, the second KL-divergence term in Equation (\ref{eqn:CD2}) will become zero. The Equation (\ref{eqn:CD2}) is reduced to minimize the KL-divergence between the observed data distribution $p_{\rm expert}$ and the energy-based model $p_{\theta}$. Eventually, $q_{\alpha}$ chases $p_{\theta}$ toward $p_{\rm expert}$.

\section{Experiments} \label{sec:experients}

We evaluate the proposed energy-based continuous inverse optimal control methods on autonomous driving tasks. The code, dataset, more results and experiment details can be found in the project page: \url{http://www.stat.ucla.edu/~yifeixu/ebm-ioc}.

\subsection{Experimental setup}

In the task of autonomous driving, the state $x_t$ consists of the coordinate, heading angle and velocity of the car, the control $u_t$ consists of steering angle and acceleration, the environment $e$ consists of road condition, speed limit, the curvature of the lane (which is represented by a cubic polynomial), as well as the coordinates of other vehicles. The trajectories of other vehicles are treated as known environment states and assumed to remain unchanged while the ego vehicle is moving, even though the trajectories of other vehicles should be predicted in reality. In this paper, we sidestep this issue and focus on the inverse optimal control problem. 
 
We assume the dynamic function of all vehicles is a non-linear bicycle model ~\cite{polack2017kinematic}, which considers longitudinal, lateral and yaw motions and assumes negligible lateral weight shift, roll and compliance steer while traveling on a smooth road. We assume all vehicles are standard two-axle, four-tire passenger cars with a 3-meter wheelbase. We set an understeering shift to be 0.043 when calculating heading~angles.

As to learning, the model parameters are randomly~initialized~by a normal distribution. The control variables are~initialized by zeros, which means keeping straight. We normalize the control variables, i.e., the steering and acceleration, because their scales are different. Instead of sampling the control variables, we sample their changes. We set the number of steps of the Langevin dynamics or the gradient descent to be $l=64$ and set the step size to be $\delta=0.2$. The choice of $l$ is a trade-off between computational efficiency and prediction accuracy. For parameter training, we use the Adam optimizer \cite{kingma2014adam}. 

We use Root Mean Square Error (RMSE) in meters with respect to each timestep $t$ to measure the accuracy of prediction, i.e., $\text{RMSE}(t) = \sqrt{\frac{1}{n}\sum_{i=1}^n \left\Vert \hat{y}_{it}- y_{it} \right\Vert^2}$, where $n$ is the number of expert demonstrations, $\hat{y}_{it}$ is the predicted coordinate of the $i$-th demonstration at time $t$ and $y_{it}$ is the ground truth coordinate of the $i$-th demonstration at timestep $t$. A small RMSE is desired. As a stochastic method, our method draws 5 samples from the learned model for prediction and the model performance is evaluated by the average RMSE and the minimum RMSE over 5 sampled trajectories. 

\subsection{Dataset}

We test our methods on two datasets. The Massachusetts driving dataset focuses on highways with curved lanes and static scenes, while the NGSIM US-101 dataset focuses on rich vehicle interactions. We randomly split each dataset into training and testing sets. The introductions of these two datasets are given below.

(1) Massachusetts driving dataset: This is a private dataset collected through a vehicle during repeated trips on a stretch of highway. The dataset includes vehicle states and controls, which are collected by the hardwares on the vehicles, as well as environment information. This dataset has a realistic driving scenario, including curved lanes and complex static scenes. To solve the problem of noisy GPS signal, Kalman filtering is used to denoise the data. There are 44,000 trajectories, each of which contains 40 0.1-second timesteps and is 4 seconds~long. 

(2) NGSIM US-101: NGSIM~\cite{ngsim} contains real highway traffics captured at 10Hz over a time span of 45 minutes. Compared to Massachusetts driving dataset, NGSIM has rich vehicle interactions. The control needs to consider other nearby vehicles. We preprocess the data by dividing the data into 5-second/50-timestep trajectories. The first 10 timesteps are for history and the remaining 40 timesteps are used for prediction. There are 831 scenes with 96,512 5-second vehicle trajectories. No control variables are provided. Thus, we need to infer the controls of each vehicle given the vehicle states. Assuming the bicycle model~\cite{polack2017kinematic} dynamics, we perform an inverse dynamics optimization using gradient descent to infer controls. In addition to minimizing the reconstruction error on states, we also minimize the L2 norm of the control variables and the difference between every two consecutive controls. The overall RMSE between the reconstructed positions and the ground truth GPS positions is 0.97 meters. The preprocessed trajectories are assumed to have perfect dynamics with noiseless and smooth sequences of controls and GPS coordinates. 

\subsection{Network structure}

We first use a linear combination of some hand-designed features as the cost function. The features include: the distance from the current vehicle to the goal point (a virtual point set at front of the vehicle) in terms of longitude and latitude, the distance to the center of the lane, the difference between the current speed and the speed limit, the difference between the vehicle direction and the lane direction, the L2 norm of the control values (including acceleration and steering), the difference between the current control value and the control value at the previous timestep (including acceleration and steering), and the distance from the vehicle to the nearest obstacle. Feature normalization is adopted to make sure that each feature has the same scale of magnitude. These features are also used to design the cost function networks of our methods, as well as baseline methods for fair comparison. 

Tables \ref{tab:appendix3} and \ref{tab:appendix4} present the multilayer perceptron (MLP) structure and the convolutional neural network (CNN) structure of cost functions, respectively, that we use in Section \ref{sec:ab}. As to the MLP structure, the number of hidden layers $N_{\text{hidden}}$ is 64 and the number of layers is 3 (i.e., 2 hidden layers and 1 output layer) by default.
The MLP cost function in Table \ref{tab:appendix3} is defined on a single frame and the cost function of the whole trajectory is the summation of costs over all frames. The CNN cost function presented in Table \ref{tab:appendix4} is defined on a trajectory with 40 time frames. Table \ref{tab:appendix2} shows the structure of the generator model used in the joint training framework. It is similar to the actor network of the PPO policy \cite{schulman2017proximal} in the generative adversarial imitation learning (GAIL) \cite{ho2016generative}. 

\begin{table}[!h]
		\caption{Network structure of the MLP cost function}
		\begin{center}
			\begin{tabular}{l|c}
				\Xhline{2\arrayrulewidth}
				Layer & Output Size \\
				\hline
				concat({[}$x, u, e${]}) & 6 + 2 + 29 \\
				hand-designed features    & 10         \\
				Linear, LeakyReLU     & $N_{\text{hidden}}$          \\
				Linear, LeakyReLU     & $N_{\text{hidden}}$          \\
				Linear                & 1       \\  
				\Xhline{2\arrayrulewidth} 
			\end{tabular}
		\end{center}
		\label{tab:appendix3}
		\vspace{-3mm}
\end{table}

\begin{table}[!h]
		\caption{Network structure of the CNN cost function}
		\begin{center}
			\begin{tabular}{l|cc}
				\Xhline{2\arrayrulewidth}
				Layer & Output Size & Stride\\
				\hline
				$\sharp$ of frames $\times$ concat({[}$x, u, e${]}) & 1 $\times$ 40 $\times$ (6 + 2 + 29) & $-$\\
				hand-designed features    & 1 $\times$ 40 $\times$ 10 &  $-$       \\
				1$\times$4 Conv1d, LeakyReLU  & 1 $\times$ 19 $\times$ 32 &    2    \\
				1$\times$4 Conv1d, LeakyReLU     & 1 $\times$ 9 $\times$ 64    & 2    \\
				1$\times$4 Conv1d, LeakyReLU     & 1 $\times$ 4 $\times$ 128   & 2 \\
				1$\times$4 Conv1d, LeakyReLU     & 1 $\times$ 1 $\times$ 256       & 1  \\
				Linear                & 1      & $-$   \\
				\Xhline{2\arrayrulewidth}
			\end{tabular}
		\end{center}
		\vspace{-3mm}
		\label{tab:appendix4}
\end{table}
\begin{table}[!h]
	\center
		\caption{Network structure of the generator model}
		\begin{tabular}{l|c}
			\Xhline{2\arrayrulewidth}
			Layer & Output Size \\	
			\hline
			concat({[}$x, e, \xi${]}) & 6 + 29 + 4 \\
			Linear, ReLU          & 64         \\
			Linear, ReLU          & 16         \\
			Linear, ReLU          & 8          \\
			Linear, Tanh          & 2          \\
			\Xhline{2\arrayrulewidth}
		\end{tabular}
		\label{tab:appendix2}
\end{table}
\begin{figure*}[t]
\centering
		\rotatebox{90}{\footnotesize\hspace{7mm} case 1}
		\fbox{%
            \includegraphics[trim=50 40 40 30,clip,width=0.184\linewidth]{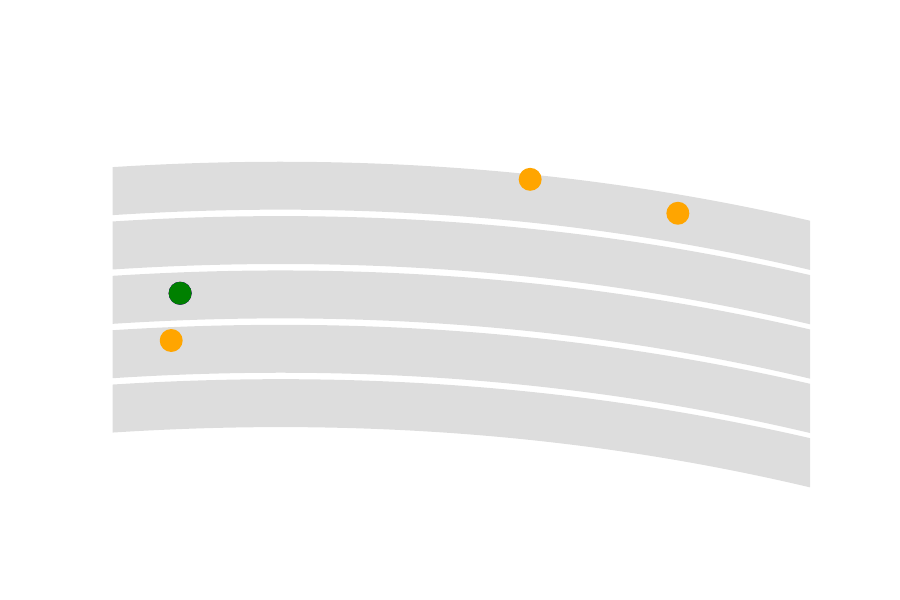}%
        }
		\fbox{%
            \includegraphics[trim=50 40 40 30,clip,width=0.184\linewidth]{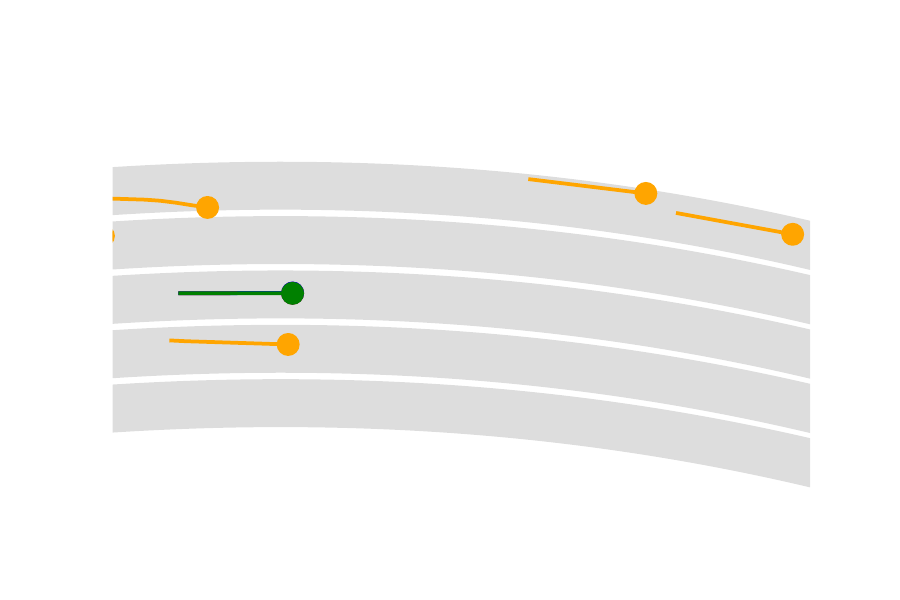}%
        }
		\fbox{%
            \includegraphics[trim=50 40 40 30,clip,width=0.184\linewidth]{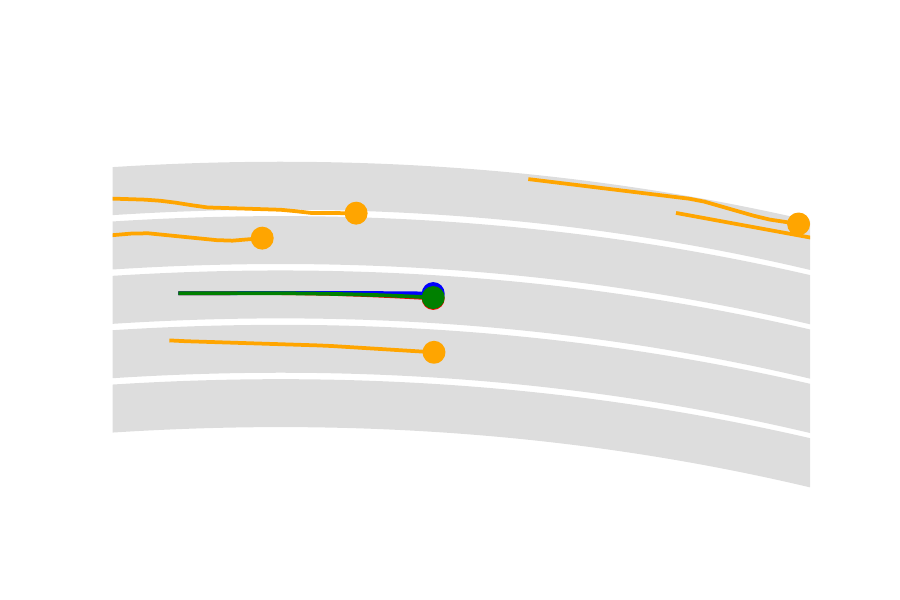}%
        }
		\fbox{%
            \includegraphics[trim=50 40 40 30,clip,width=0.184\linewidth]{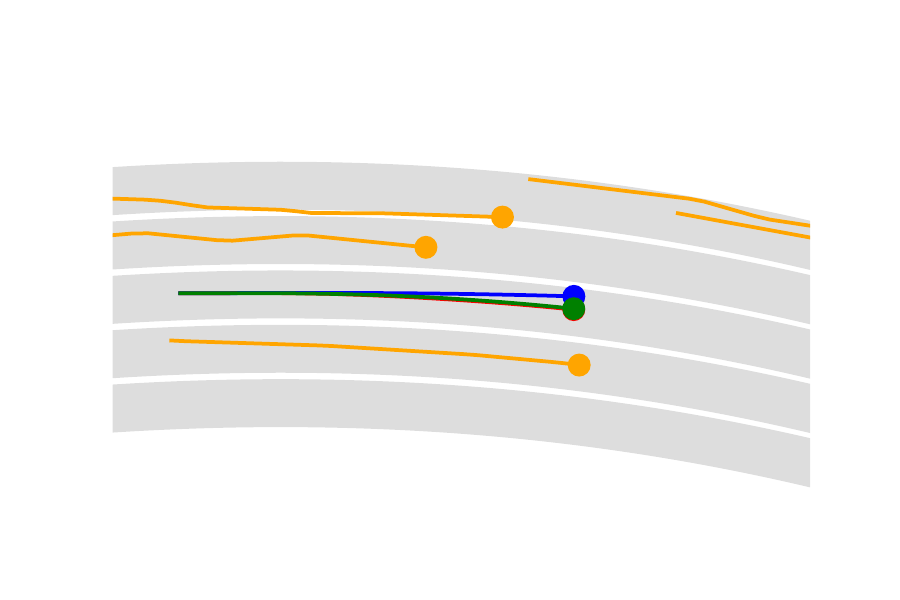}%
        }
		\fbox{%
            \includegraphics[trim=50 40 40 30,clip,width=0.184\linewidth]{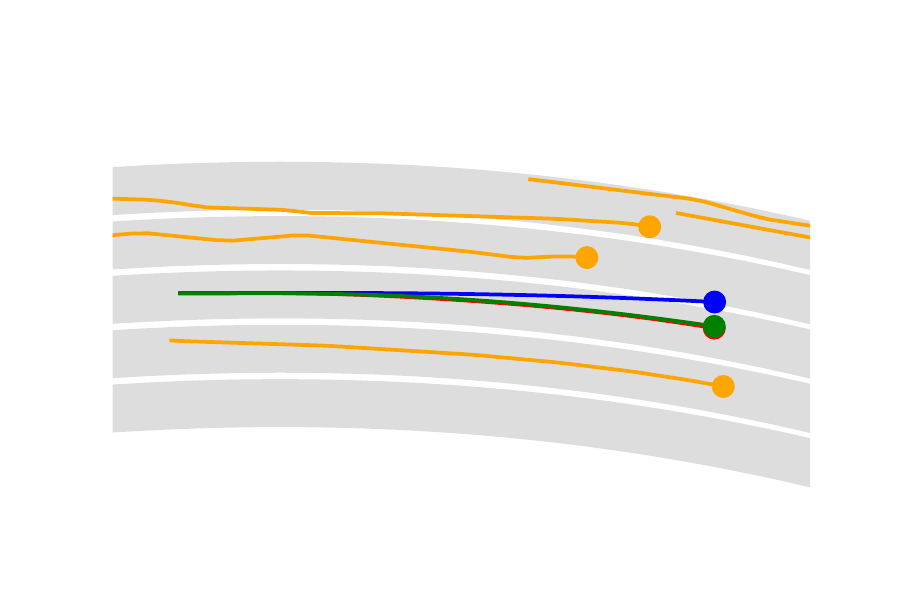}%
        }\\
        \vspace{1.4mm}
		\rotatebox{90}{\footnotesize\hspace{7mm} case 2}
		\fbox{%
            \includegraphics[trim=50 40 40 30,clip,width=0.184\linewidth]{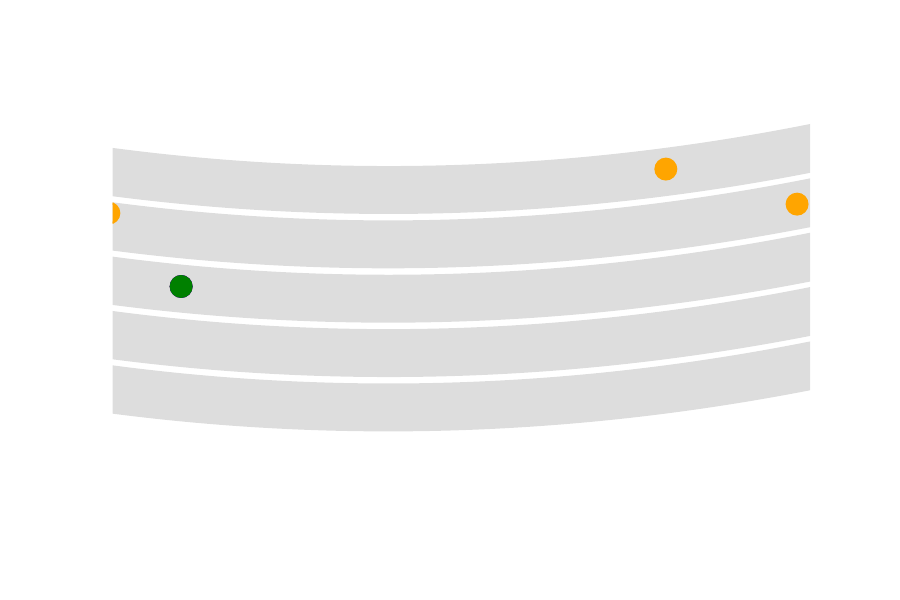}%
        }
		\fbox{%
            \includegraphics[trim=50 40 40 30,clip,width=0.184\linewidth]{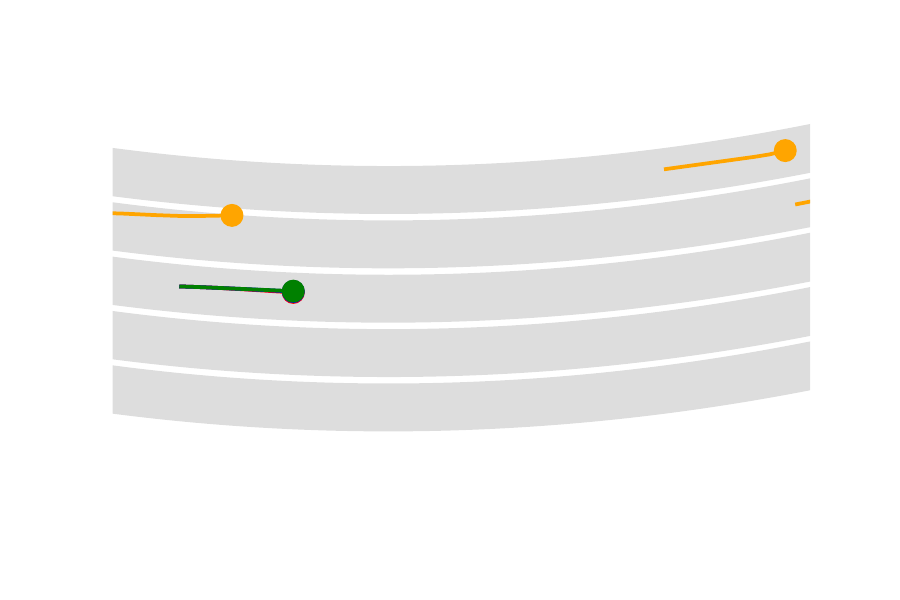}%
        }
		\fbox{%
            \includegraphics[trim=50 40 40 30,clip,width=0.184\linewidth]{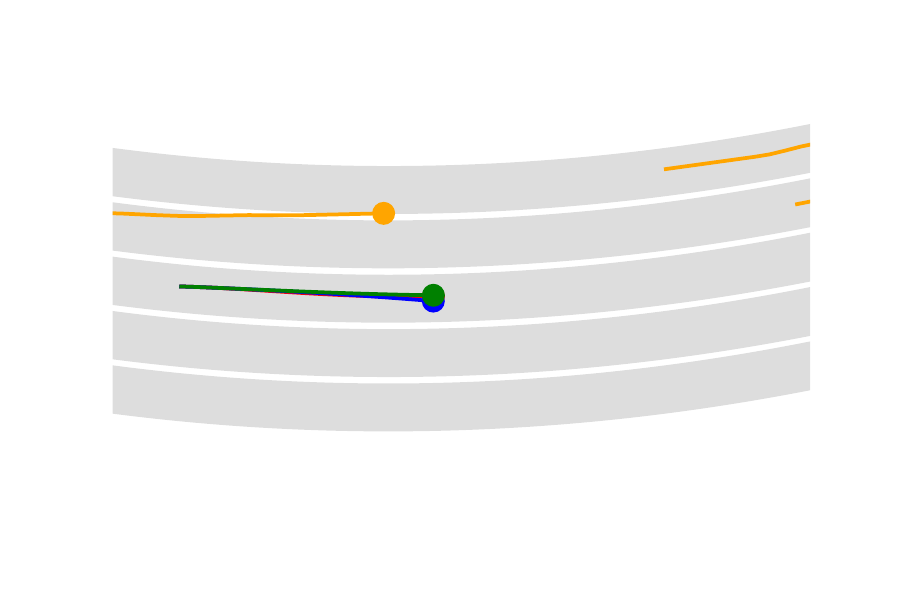}%
        }
		\fbox{%
            \includegraphics[trim=50 40 40 30,clip,width=0.184\linewidth]{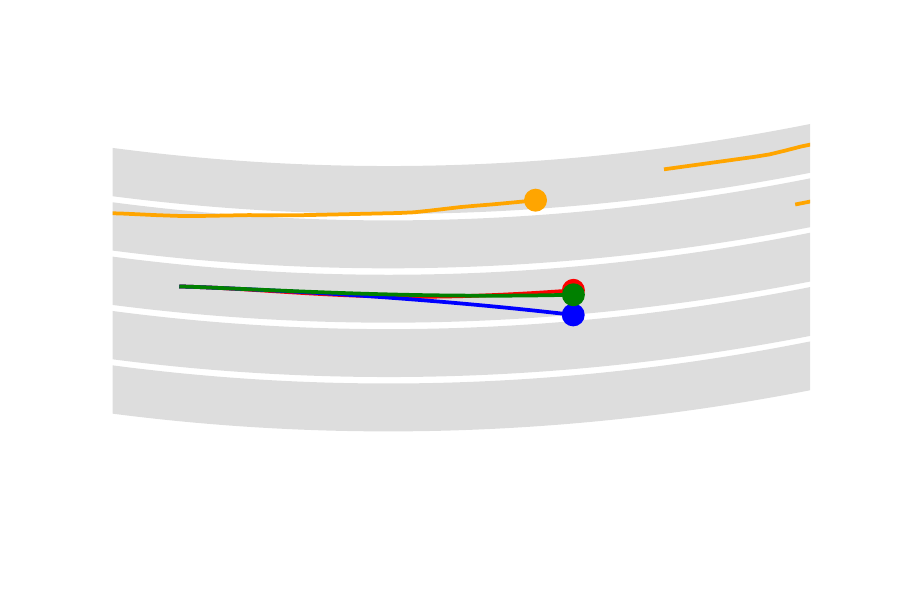}%
        }
		\fbox{%
            \includegraphics[trim=50 40 40 30,clip,width=0.184\linewidth]{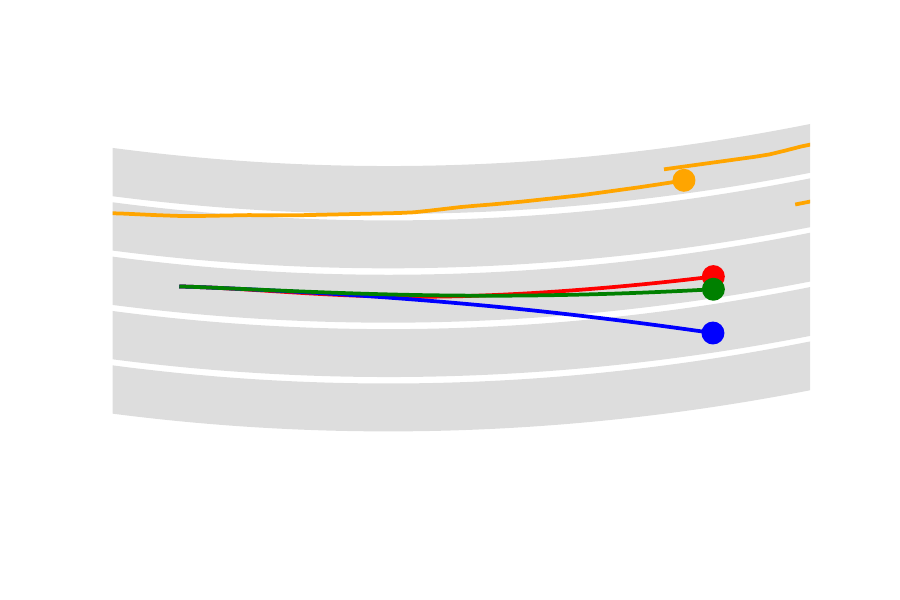}%
        } \\
{\footnotesize \hspace{12mm}  starting point \hspace{20mm} 1st second \hspace{20mm} 2nd second \hspace{20mm} 3rd second  \hspace{20mm} 4th second \hspace{10mm}}
\caption{Predicted trajectories for single-agent control on the Massachusetts driving dataset. The staring point is the last frame of the history trajectory. (Green: predicted trajectories by our model. Blue: predicted trajectories by GAIL. Red: ground truth trajectories. Orange: trajectories of other vehicles. Gray: lanes.)}\label{fig:isee_singles}
\end{figure*}
\subsection{Training details}

\textbf{Normalization}. We apply normalization to the controls (i.e., acceleration and steering) and hand-designed features. For the controls, we normalize their values to have zero mean and unit variance. We also normalize each hand-designed feature by dividing  by the mean. We normalize two datasets separately.

\textbf{Optimizer}. We use the Adam optimizer on training our models. Both $\beta_1$ and $\beta_2$ are set to be 0.5. All model parameters are randomly initialized by the He initialization method \cite{HeZRS15}, which is a uniform distribution. In the linear setting, we set learning rate of the Adam to be 0.1 with an exponential decay rate 0.999. For the MLP cost function setting, we set the learning rate to be 0.005 without an exponential decay. In the CNN cost function setting, we set the learning rate to be 0.005 with an exponential decay rate 0.999. For the training of the generator model, the learning rate is 0.002 and the exponential decay rate is 0.998. For each epoch, we shuffle the whole dataset. The batch size is 1,024.

\textbf{Langevin Dynamics}. 
To prevent gradient values from being too large in each Langevin step, we set the maximum limit to be 0.1. The Langevin step size is set to be 0.1 and the number of Langevin steps is 64. All settings are the same for the gradient descent method to synthesize the controls. %In appendix 3, we do an ablation study on the different 

\textbf{iLQR} As to the iLQR solver, we perform a grid search for the learning rate from 0.001 to 1. The maximum step is 100. If the difference between the current step and the previous step is smaller than 0.001, early stop is triggered. In the experiment, the average number of iLQR steps is around 30.

\subsection{Single-agent control}

We first test our methods, including sampling-based and optimization-based ones, on a single-agent control problem. We compare our method with three baseline methods below:
\begin{itemize}
	\item Constant velocity: the simplest baseline with a constant velocity and a zero steering.
	\item Generative adversarial imitation learning (GAIL) \cite{ho2016generative}: The original GAIL method was proposed for imitation learning. We use the same setting as in~\cite{kuefler2017imitating}, which applies the GAIL to the task of modeling human highway driving behavior. Besides, we change the policy gradient method from Trust Region Policy Optimization (TRPO) \cite{schulman2015trust} to Proximal Policy Optimization (PPO) \cite{schulman2017proximal}. 
	\item IOC with Laplace~\cite{levine2012continuous} (IOC-Laplace): We implement this baseline with the same iLQR method as that in our model. 
\end{itemize}

It takes roughly 0.1 seconds to predict a full trajectory with a 64-step Langevin dynamics (or gradient descent). Figure \ref{fig:isee_singles} displays two qualitative results. Each row shows one 4-frame example with a frame interval equal to 1 second. Each frame shows trajectories over time for different vehicles as well as different baseline methods for comparison. Table \ref{tab:isee_number} and \ref{tab:ngsim_number} show quantitative results for Massachusetts driving dataset and NGSIM, respectively. In the last two rows, we provide both average RMSE and minimum RMSE for our sampling-based approach. Our methods achieve substantial improvements compared to baseline methods, such as IOC-Laplace \cite{levine2012continuous} and GAIL, in terms of testing RMSE. We find that the sampling-based methods outperform the optimization-based methods among our energy-based approaches. 

\begin{table}[h]
	\begin{center}
		\caption{Massachusetts driving dataset results (RMSE).}\label{tab:isee_number}
		\begin{tabular*}{\hsize-20pt}{@{\extracolsep{\fill}}l|ccc}
			\Xhline{2\arrayrulewidth}
			Method    & 1s    & 2s    & 3s   \\
			\Xhline{1\arrayrulewidth}
			Constant Velocity & 0.340 & 0.544 & 1.023\\
			IOC-Laplace & 0.386 & 0.617 & 0.987\\
			GAIL & 0.368 & 0.626 & 0.977\\
			ours (via iLQR) & 0.307 & 0.491 & 0.786\\
			ours (via GD)  & 0.257 & 0.413 & 0.660\\
			ours AVG (via Langevin)  & \textbf{0.255} & \textbf{0.401} & \textbf{0.637}\\
			ours MIN (via Langevin)  & \textit{0.157} & \textit{0.354} & \textit{0.607}\\
			\Xhline{2\arrayrulewidth}
		\end{tabular*}
	\end{center}
\end{table}
\begin{table}[h]
	\begin{center}
		\caption{NGSIM dataset results (RMSE).}\label{tab:ngsim_number}
		\begin{tabular*}{\hsize}{@{\extracolsep{\fill}}l|cccc}
			\Xhline{2\arrayrulewidth}
			Method     & 1s    & 2s    & 3s    & 4s    \\
			\hline
			Constant Velocity           & 0.569 & 1.623 & 3.075 & 4.919 \\
			IOC-Laplace & 0.503 & 1.468 & 2.801 & 4.530 \\
			GAIL                & 0.367  & 0.738  & 1.275  & 2.360  \\
			ours (via iLQR)              & 0.351 & 0.603 & 0.969 & 1.874 \\
			ours (via GD)          & 0.318 & 0.644 & 1.149 & 2.138 \\
			ours AVG (via Langevin)          & \textbf{0.311} & \textbf{0.575} & \textbf{0.880} & \textbf{1.860} \\
			ours MIN (via Langevin)  & \textit{0.203} & \textit{0.458} & \textit{0.801} & \textit{1.690}\\
			\Xhline{2\arrayrulewidth}
		\end{tabular*}
	\end{center} 
	\label{tab:classification}
\end{table}

The reason why the method ``IOC-Laplace'' performs poorly on both two datasets is due to the fact that its Laplace approximation is not accurate enough for a complex cost function used in the current tasks. 
Our models are more genetic and do not make such an approximation. Instead, they use Langevin sampling for maximum likelihood training. Therefore, they can provide more accurate prediction results.   

The problem of GAIL is its model complexity. GAIL parameterizes its discriminator, policy and value function by MLPs. Designing optimal MLP structures of these three components for GAIL is challenging. Our method only needs to design a single architecture for the cost function. 

Additionally, the optimal control of our method is performed by simulating trajectories of actions and states according to the learned cost function that takes into account the future information. In contrast, the GAIL relies on its learned policy net for step-wise decision making.

Compared with gradient descent (optimization-based approach), Langevin dynamics-based method can obtain smaller errors. One reason is that the sampling-based approach rigorously maximizes the log-likelihood of the expert demonstrations during training, while the optimization-based approach is just a convenient approximation. The other reason is that the Gaussian noise term in each Langevin step helps to explore the cost function and avoid sub-optimal solutions. 

\subsection{Corner case testing with toy examples}

\begin{figure}[h]
	\centering
	\fbox{% 
        \includegraphics[trim=60 40 50 50,clip,width=0.32\linewidth]{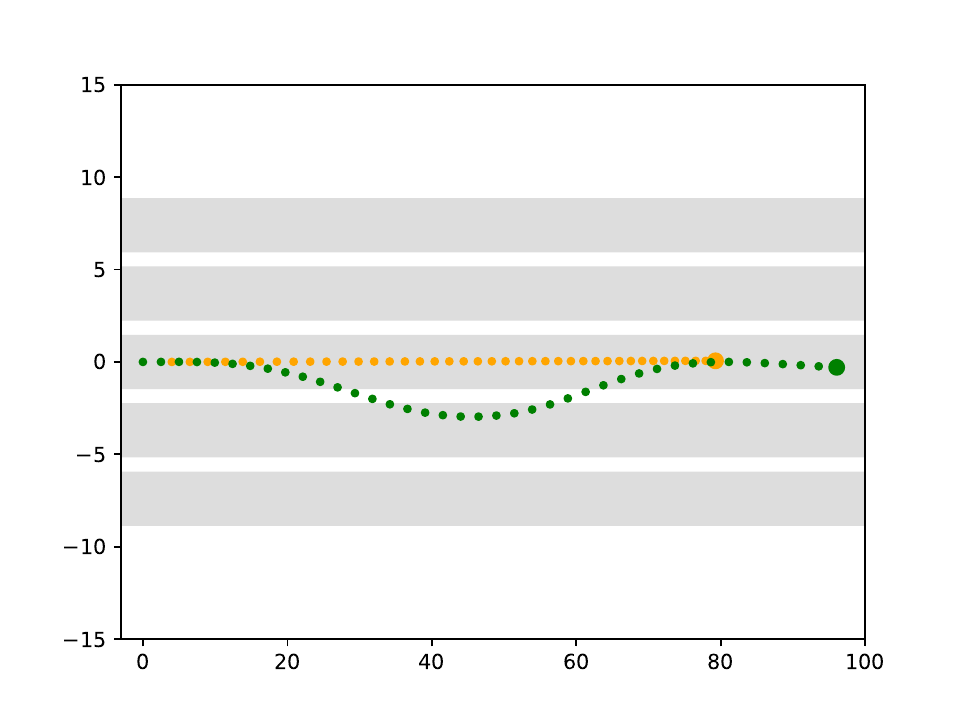}%
    }
	\fbox{% 
        \includegraphics[trim=60 40 50 50,clip,width=0.32\linewidth]{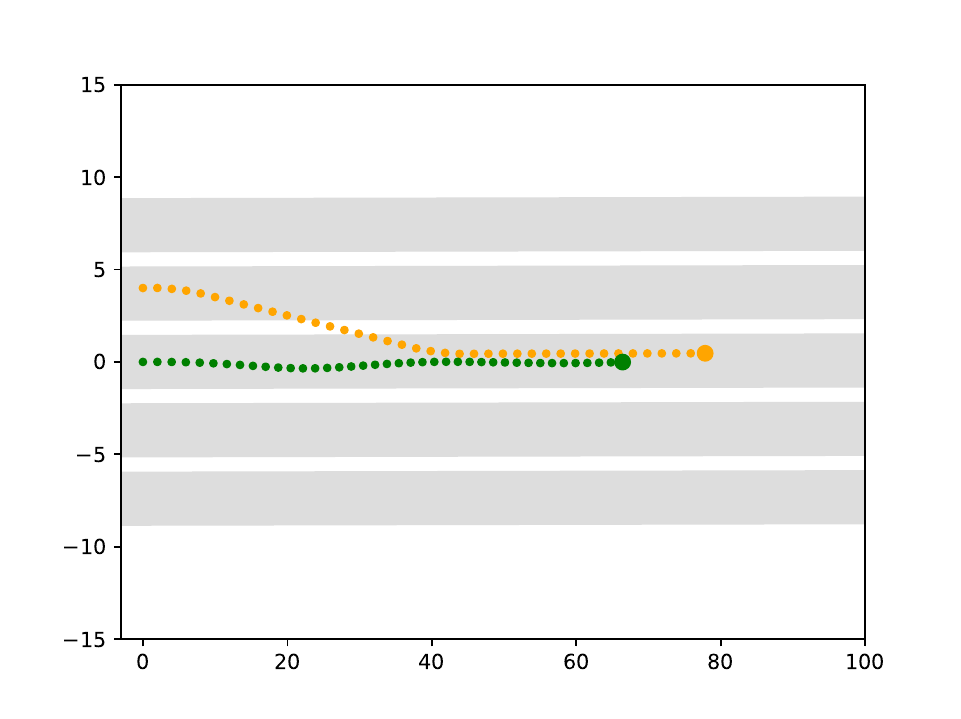}%
    }
	\fbox{% 
        \includegraphics[trim=60 40 50 50,clip,width=0.32\linewidth]{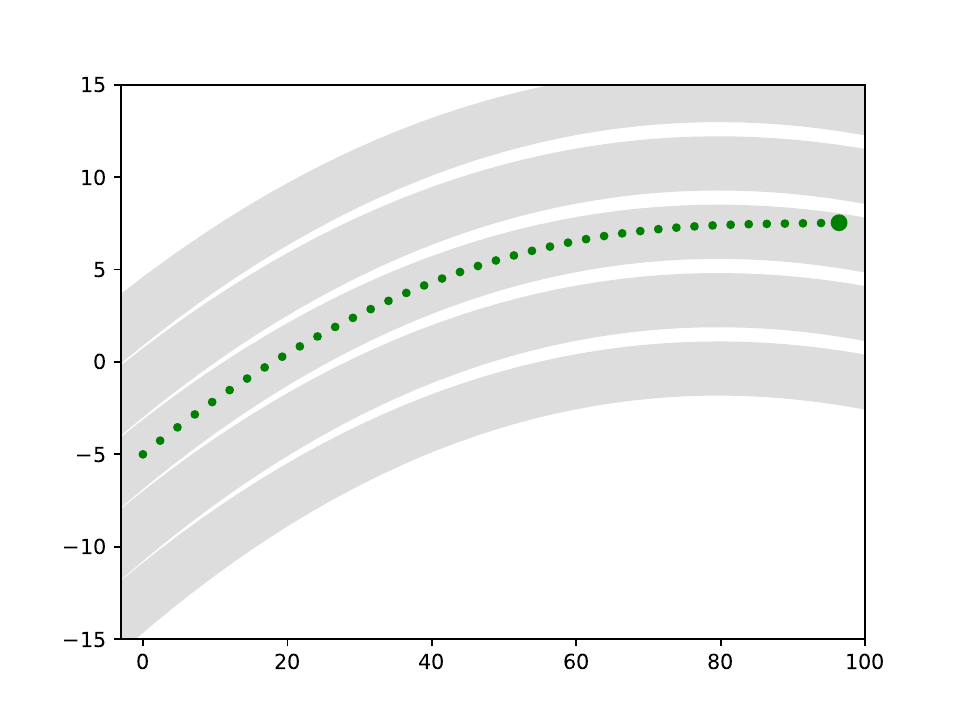}%
    } \\
	\vspace{-0.5mm}
	{\scriptsize (a)\hspace{75pt}(c)\hspace{75pt}(e)\hspace{80pt}} \\
	\vspace{1mm}
    \fbox{% 
            \includegraphics[trim=60 40 50 50,clip,width=0.32\linewidth]{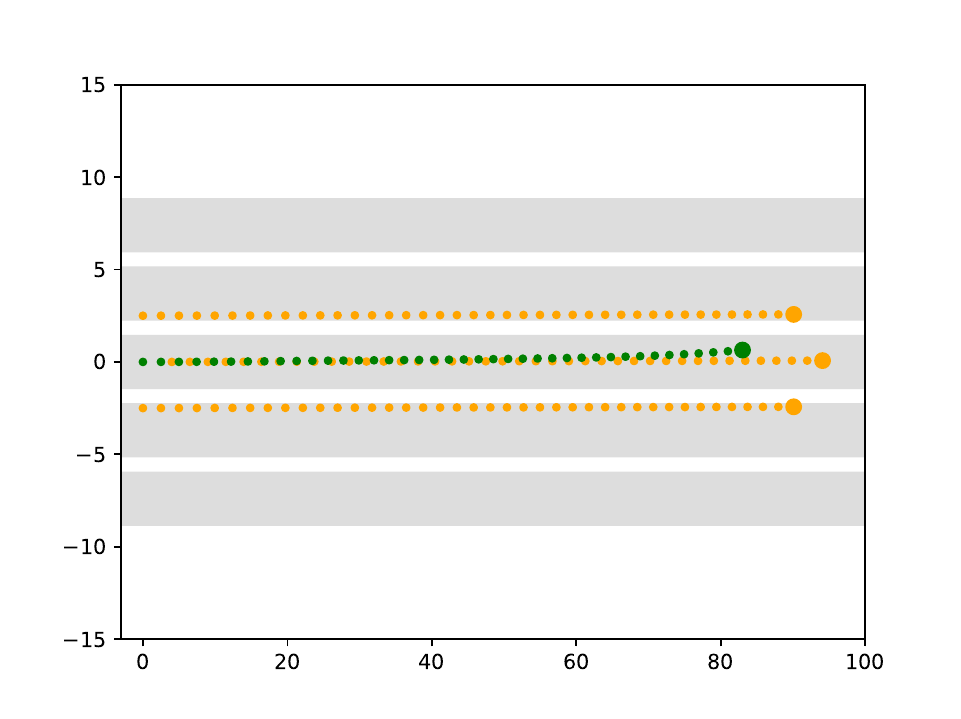}%
        }
    	\fbox{% 
            \includegraphics[trim=60 40 50 50,clip,width=0.32\linewidth]{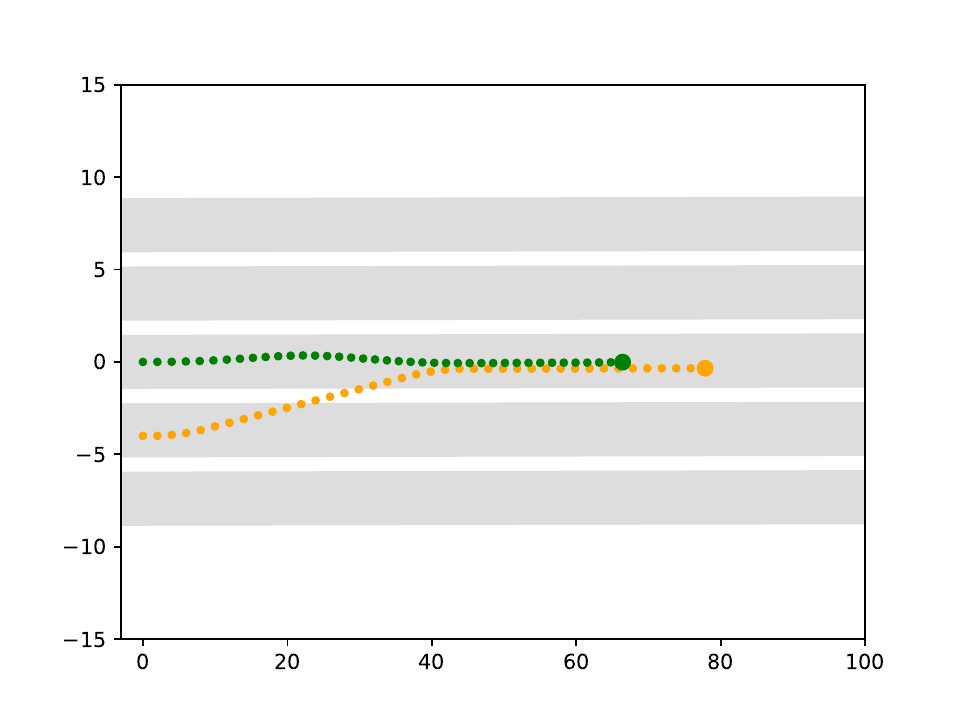}%
        }
	\fbox{% 
        \includegraphics[trim=60 40 50 50,clip,width=0.32\linewidth]{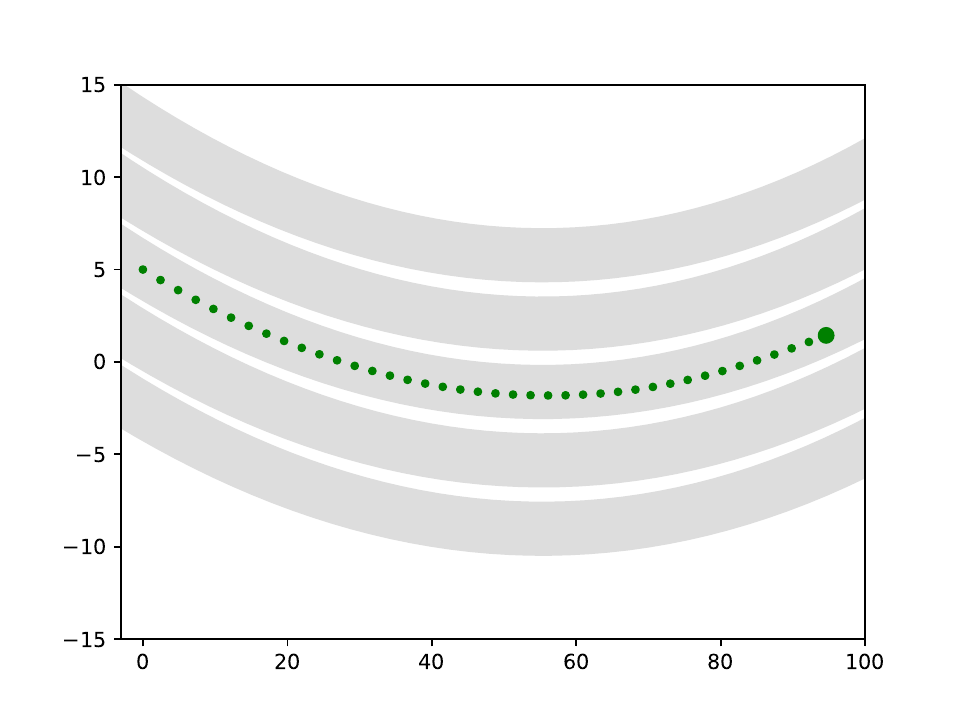}%
    } \\
	\vspace{-0.5mm}
	{\scriptsize (b)\hspace{75pt}(d)\hspace{75pt}(f)}\\
	{\scriptsize Sudden braking \hspace{40pt}Lane cut-in\hspace{30pt} Large lane curvature}
	\caption{ Prediction in corner cases. (Green : predicted trajectories. Orange : trajectories of other vehicles. Gray: lanes.)}\label{fig:toy_traj}
\end{figure}

Corner cases are important for model evaluation. We construct 6 typical corner cases to test our model. Figure \ref{fig:toy_traj} shows the predicted trajectories by our method for several cases. 
Figures \ref{fig:toy_traj}(a) and \ref{fig:toy_traj}(b) show two cases of the sudden braking. In each of the cases, a vehicle (orange) in front of the ego vehicle (green) is making a sudden brake. In case (a), there are not any other vehicles moving alongside the ego vehicle, so it is predicted to firstly change the lane, then accelerate past the vehicle in front, and return to its previous lane and continue its driving. In case (b), two vehicles are moving alongside the ego vehicle. The predicted trajectory shows that the ego vehicle is going to trigger a brake to avoid a potential collision accident. Figures \ref{fig:toy_traj}(c) and \ref{fig:toy_traj}(d) show two cases in the cut-in situation. In each case, a vehicle is trying to cut in from the left or right lane. The ego vehicle is predicted to slow down to ensure the safe cut-in of the other vehicle. Figures \ref{fig:toy_traj}(e) and \ref{fig:toy_traj}(f) show two cases in the large lane curvature situation, where our model can still perform well to predict reasonable trajectories.

\begin{figure}[h]
 	\centering
 	\includegraphics[trim=30 20 30 30,clip,width=0.32\linewidth]{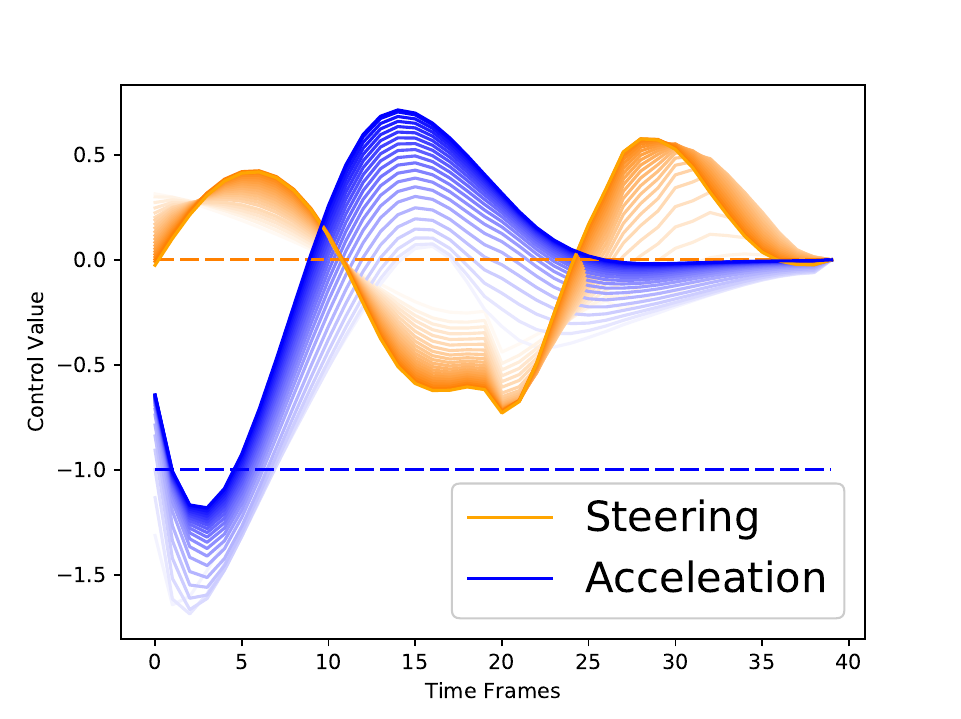}
 	\includegraphics[trim=30 20 30 30,clip,width=0.32\linewidth]{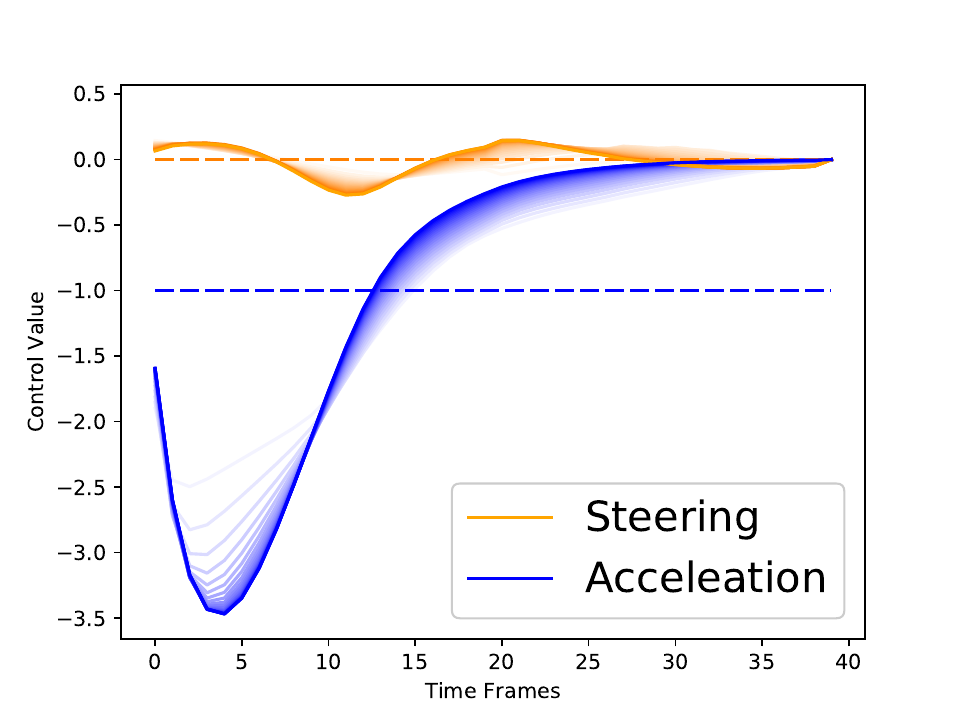} 	
 	\includegraphics[trim=30 20 30 30,clip,width=0.32\linewidth]{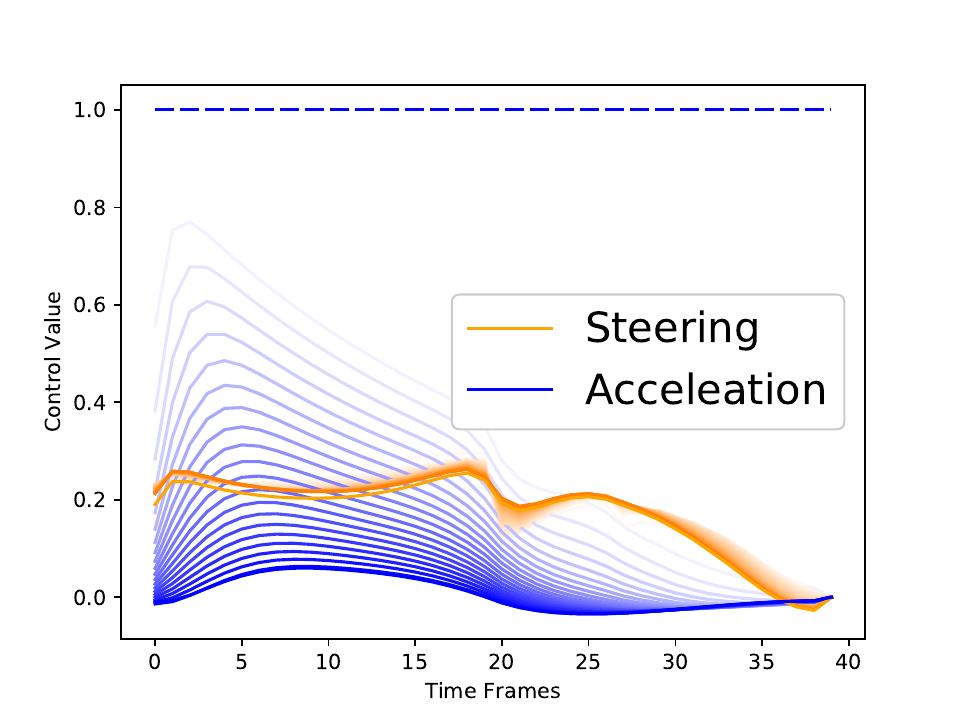}
 	{\scriptsize \hspace{50pt} (a)\hspace{70pt}(c)\hspace{70pt}(e)\hspace{60pt}}
 	\includegraphics[trim=30 20 30 30,clip,width=0.32\linewidth]{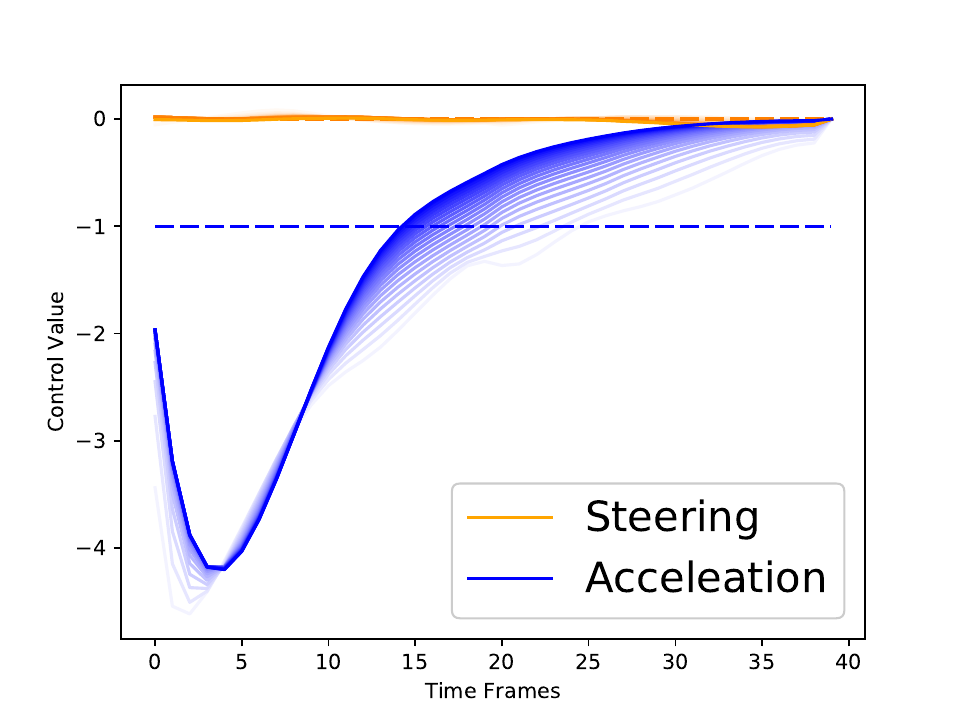}
 	\includegraphics[trim=30 20 30 30,clip,width=0.32\linewidth]{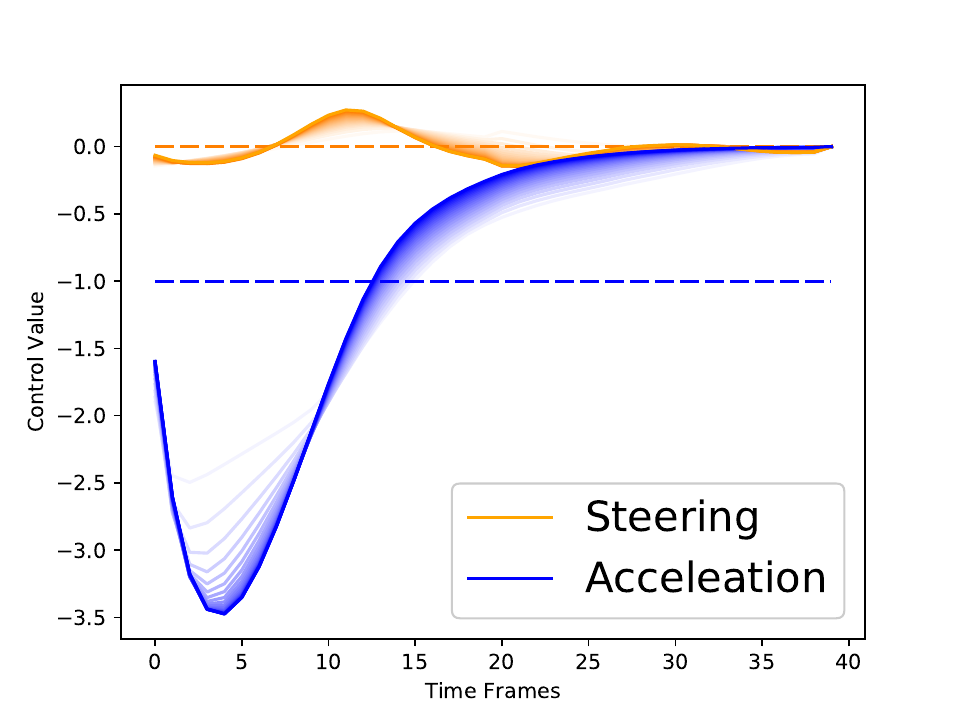} 	
 	\includegraphics[trim=30 20 30 30,clip,width=0.32\linewidth]{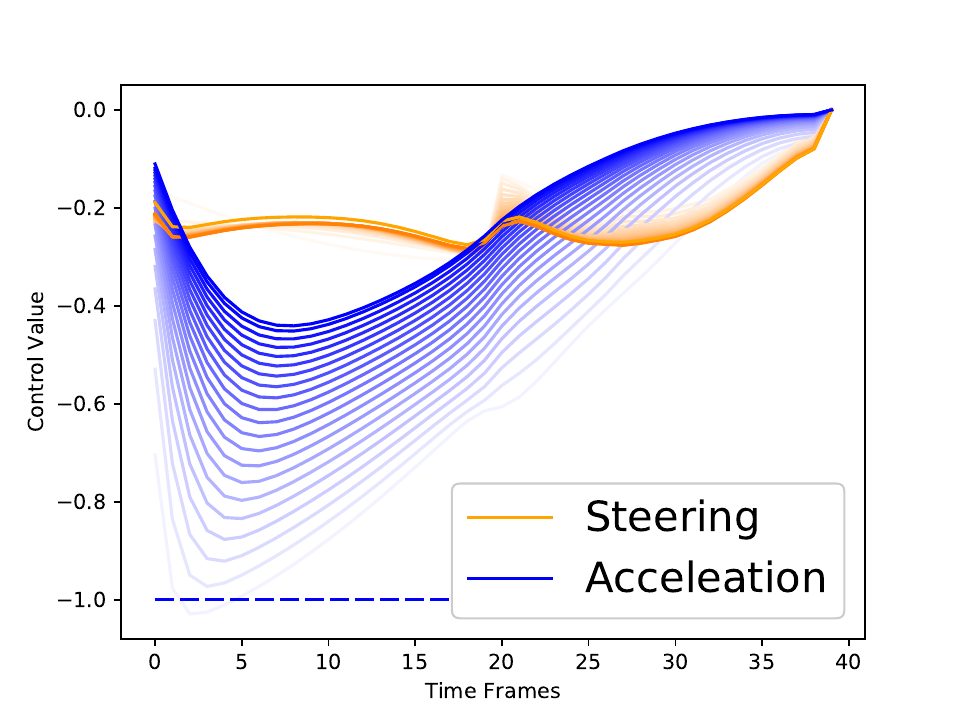}
 	{\scriptsize \hspace{50pt} (b)\hspace{70pt}(d)\hspace{70pt}(f)}\\
 	{\scriptsize Sudden braking \hspace{40pt}Lane cut-in\hspace{30pt} Large lane curvature}
 	\caption{Predicted controls over time. (Dash lines: initial values of the controls for Langevin sampling. Solid lines: predicted controls over time steps. Blue: control of acceleration. Orange: control of steering.)}\label{fig:toy_control}
 \end{figure}

Figure \ref{fig:toy_control} shows the corresponding plots of the predicted controls, i.e., steering and acceleration,  over time steps. In each plot, blue lines stand for acceleration and orange lines stand for steering. The dash lines represent the initialization of the controls for Langevin sampling, which are actually the controls at the last time steps of the history trajectories. We use 64 Langevin steps to sample the controls from the learned cost function. We plot the predicted controls (i.e., acceleration and steering) over time for each Langevin step. The curves with more numbers of Langevin steps appear darker. Thus, the darkest solid lines are the final predicted~trajectories~of~controls. 

In short, this experiment demonstrates that our method is capable of learning a reasonable cost function that handles corner cases, such as situations of sudden braking, lane cut-in, and making turns in curved lanes.


\subsection{Evaluation of different cost functions}
\label{sec:ab}
The neural network is a powerful function approximator. It is capable of approximating any complex nonlinear function given sufficient training data, and it is also flexible to incorporate prior information, which in our case are the manually designed features. In this experiment, we replace the linear cost function in our sampling-based approach with a neural network cost function. Specifically, we design a cost function by multilayer perceptron (MLP), where we put three layers on top of the vector of hand-designed features: $C_\theta({\bf x}, {\bf u}, e, h) = f(\phi({\bf x}, {\bf u}, e, h))$, where $f$ contains 2 hidden layers and 1 output layer, and $\theta$ contains all trainable parameters in $f$. We also consider using a 1D CNN that takes into account the temporal relationship inside the trajectory for the cost function. We add four 1D convolutional layers on top of the sequence of vectors of hand-designed features, where the kernel size in each layer is 1 $\times$ 4. The numbers of channels are $\{32, 64, 128, 256\}$ and the numbers of strides are $\{2, 2, 2, 1\}$ for different layers, respectively. One fully connected layer with a single kernel is attached at the end.

Table \ref{tab:mlp_cost_res} shows a comparison of performances of different designs of cost functions. We can see that improvements can be obtained by using cost functions parameterized by either MLP or CNN. Neural network provides nonlinear connection layers as a transformation of the original input features. This implies that there are some internal connections between the features and some temporal connections among feature vectors at different time steps.
\begin{table}[h]
\begin{center}
	\caption{A comparison of performances of models with different cost functions (average RMSE).}\label{tab:mlp_cost_res}
	\begin{tabular*}{\hsize-20pt}{@{\extracolsep{\fill}}l|ccc}
		\Xhline{2\arrayrulewidth}
		Method   & 1s    & 2s    & 3s   \\
		\hline
		Linear cost function & 0.255 & 0.401 & 0.637\\
		MLP cost function & 0.237 & 0.379 & 0.607\\
		CNN cost function & \textbf{0.234} & \textbf{0.372} & \textbf{0.572}\\
		\Xhline{2\arrayrulewidth}
	\end{tabular*}
\end{center}
\end{table}

\begin{figure*}[h]
\centering
		\rotatebox{90}{\footnotesize\hspace{8mm} case 1}
		\fbox{%
		    \includegraphics[trim=60 40 50 50,clip,width=0.184\linewidth,page=1]{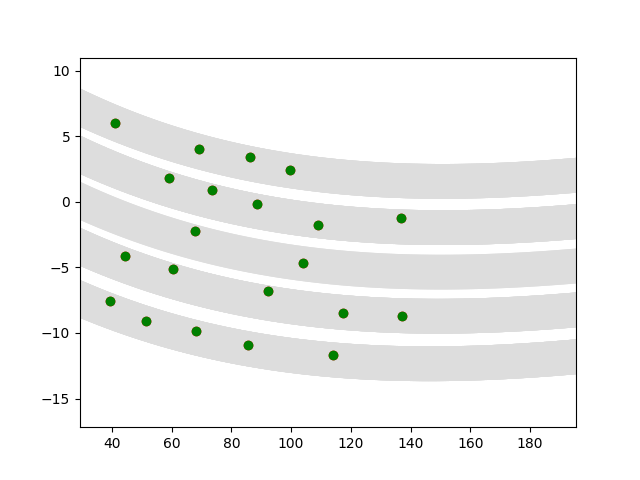}%
		}
		\fbox{%
            \includegraphics[trim=60 40 50 50,clip,width=0.184\linewidth]{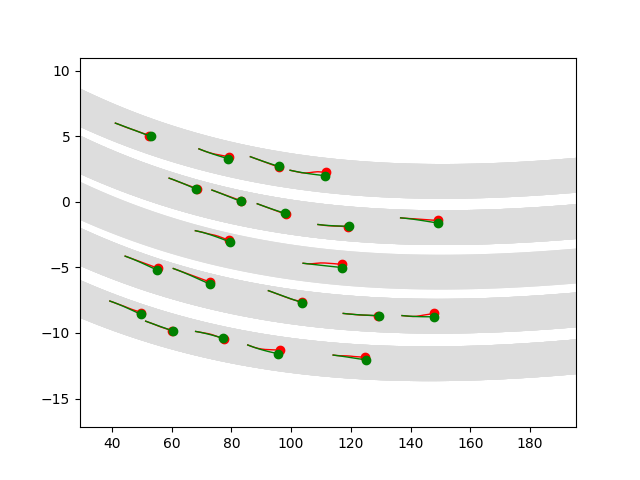}%
        }
		\fbox{%
            \includegraphics[trim=60 40 50 50,clip,width=0.184\linewidth]{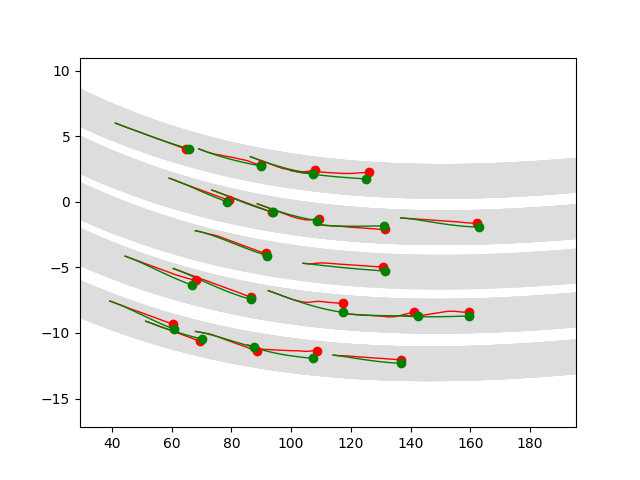}%
        }
		\fbox{%
            \includegraphics[trim=60 40 50 50,clip,width=0.184\linewidth]{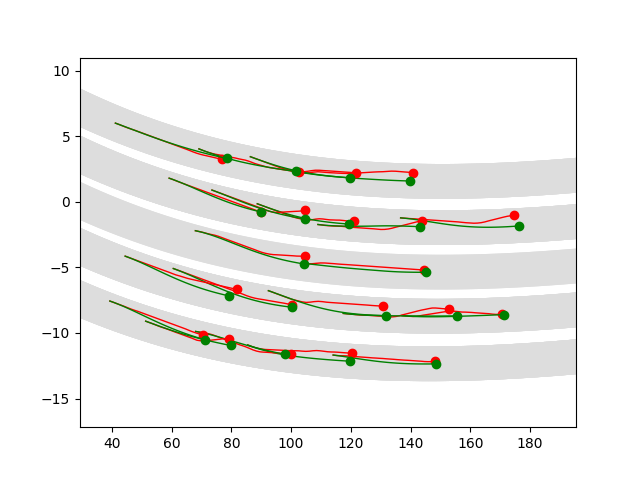}%
        }
		\fbox{%
            \includegraphics[trim=60 40 50 50,clip,width=0.184\linewidth]{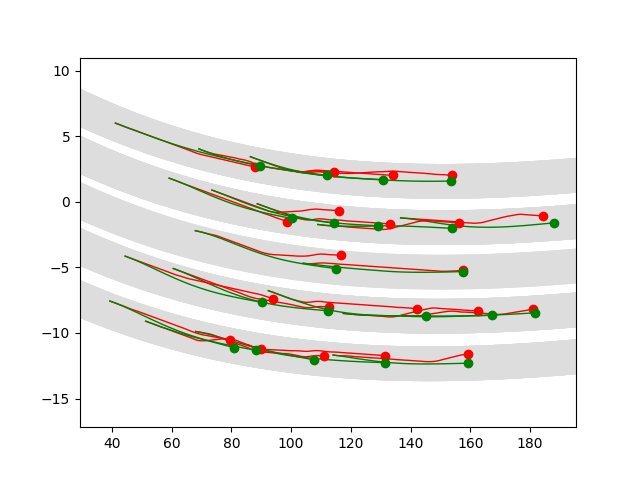}%
        }\\
        \vspace{1.4mm}
		\rotatebox{90}{\footnotesize\hspace{8mm} case 2}
		\fbox{%
            \includegraphics[trim=60 40 50 50,clip,width=0.184\linewidth]{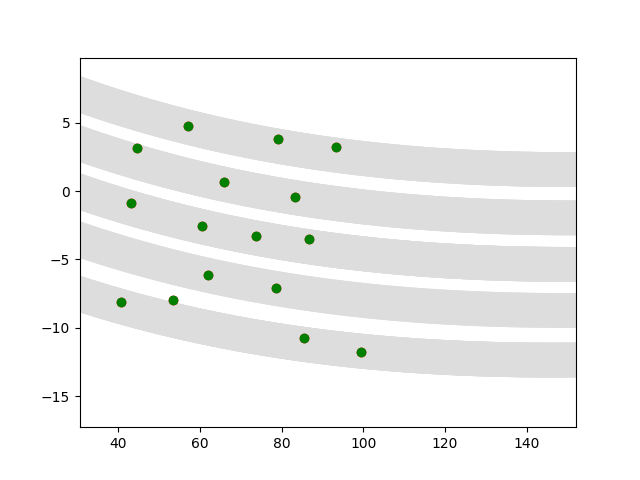}%
        }
		\fbox{%
            \includegraphics[trim=60 40 50 50,clip,width=0.184\linewidth]{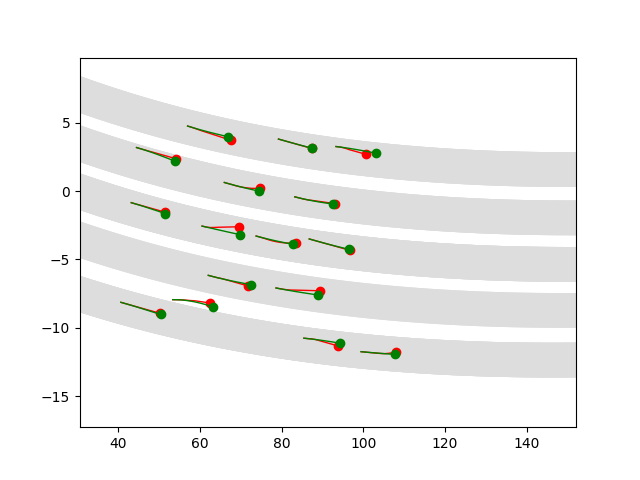}%
        }
		\fbox{%
            \includegraphics[trim=60 40 50 50,clip,width=0.184\linewidth]{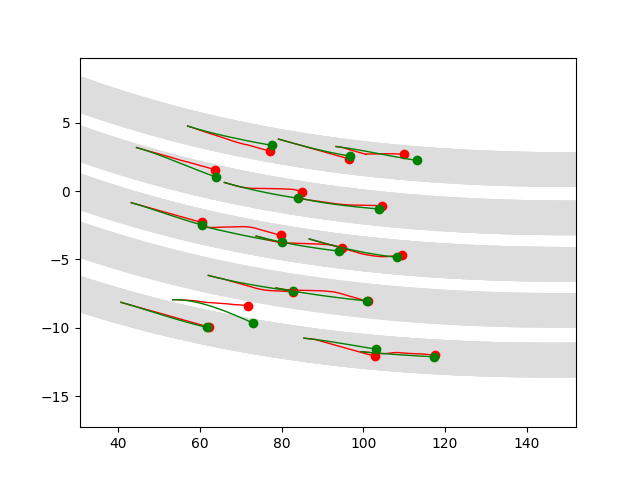}%
        }
		\fbox{%
            \includegraphics[trim=60 40 50 50,clip,width=0.184\linewidth]{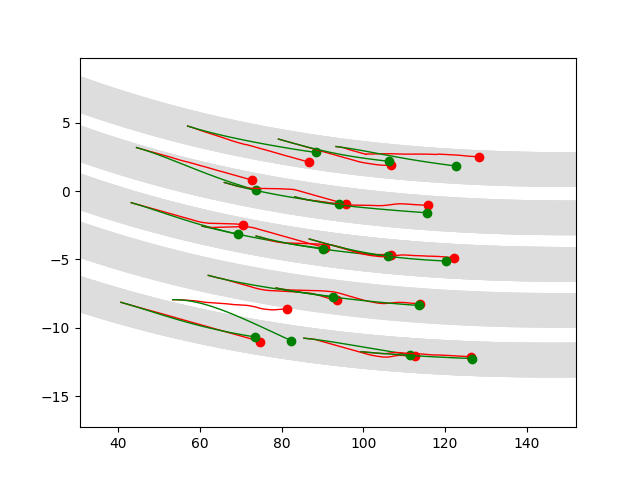}%
        }
		\fbox{%
            \includegraphics[trim=60 40 50 50,clip,width=0.184\linewidth]{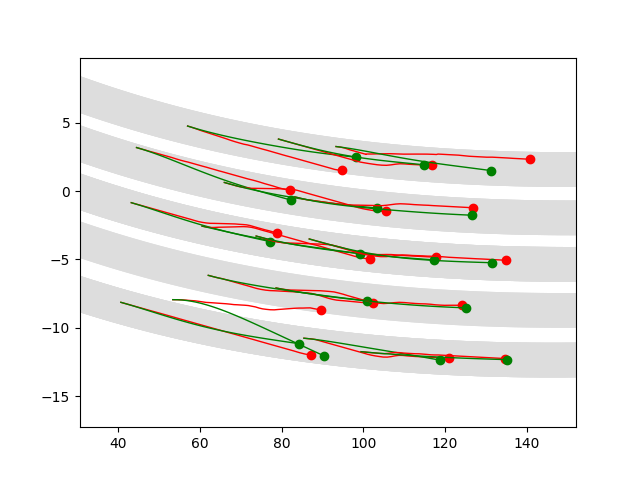}%
        } \\
{\footnotesize \hspace{12mm}  starting point \hspace{20mm} 1st second \hspace{20mm} 2nd second \hspace{20mm} 3rd second  \hspace{20mm} 4th second \hspace{10mm}}
		\caption{Predicted trajectories for multi-agent control on the NGSIM dataset. The starting point is the last frame of the history trajectory. (Green: predicted trajectories by our model. Red: ground truth trajectories. Gray: lanes.)}\label{fig:multi_traj} 
\end{figure*}

\subsection{Multi-agent control}

In the setting of single-agent control, the future trajectories of other vehicles are assumed to be given and they remain unchanged no matter how the ego vehicle moves. We extend our energy-based framework to the multi-agent setting, in which we simultaneously control all vehicles in the scene. The controls of other vehicles are used to predict the trajectories of other vehicles. 

Suppose there are $K$ agents, and every agent in the scene can be regarded as a general agent. The state and control space are Cartesian products of the individual states and controls respectively, i.e., ${\bf X} = ({\bf x}^k , k=1, 2, ..., K), {\bf U} = ({\bf u}^k , k=1, 2, ..., K)$. All the agents share the same dynamic function, which is $x_{t}^k = f(x_{t-1}^k, u_t^k), \forall k=1, 2, ..., K$. The overall cost function are set to be the sum of each agent $C_\theta({\bf X}, {\bf U}, e, h) = \sum_{k=1}^K{C_\theta({\bf x}^k, {\bf u}^k, e, h^k)}$. Thus, the conditional probability density function becomes $p_\theta({\bf U}|e, h) = \frac{1}{Z_\theta(e, h)} \exp[ - C_\theta({\bf X}, {\bf U}, e, h) ]$, where $Z_\theta(e, h)$ is the intractable normalizing constant.

\begin{figure*}[h]
	\centering
	\includegraphics[width=0.26\linewidth]{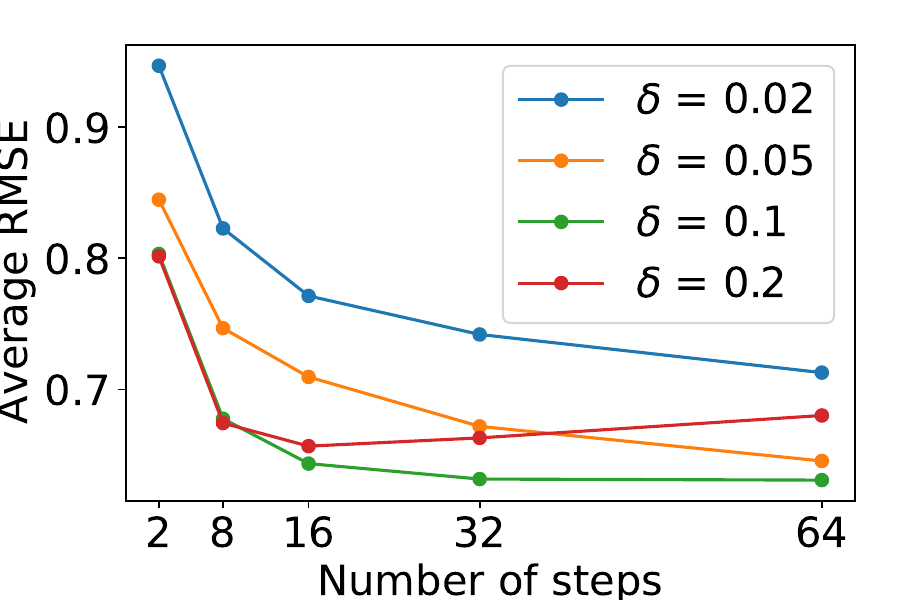}
	\includegraphics[width=0.26\linewidth]{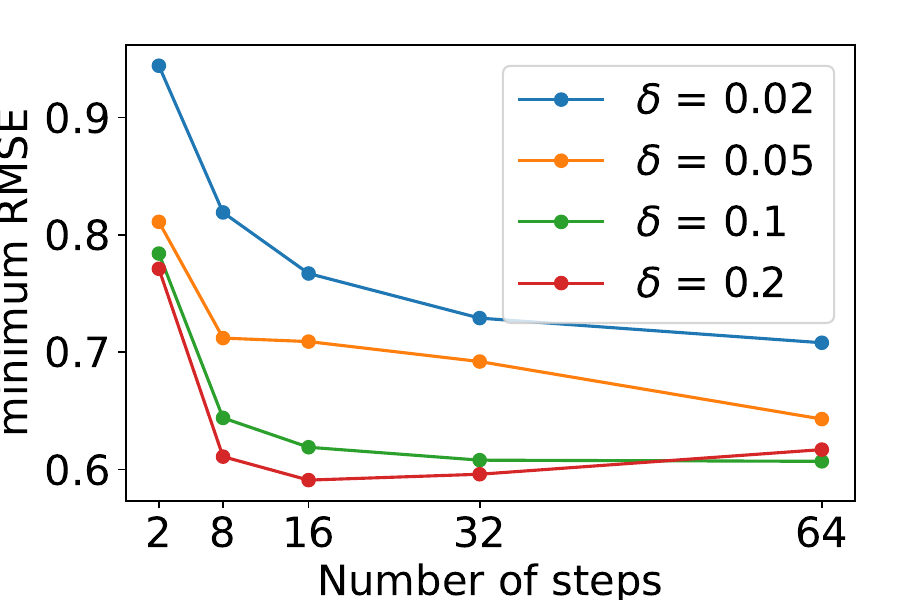}
	\includegraphics[width=0.26\linewidth]{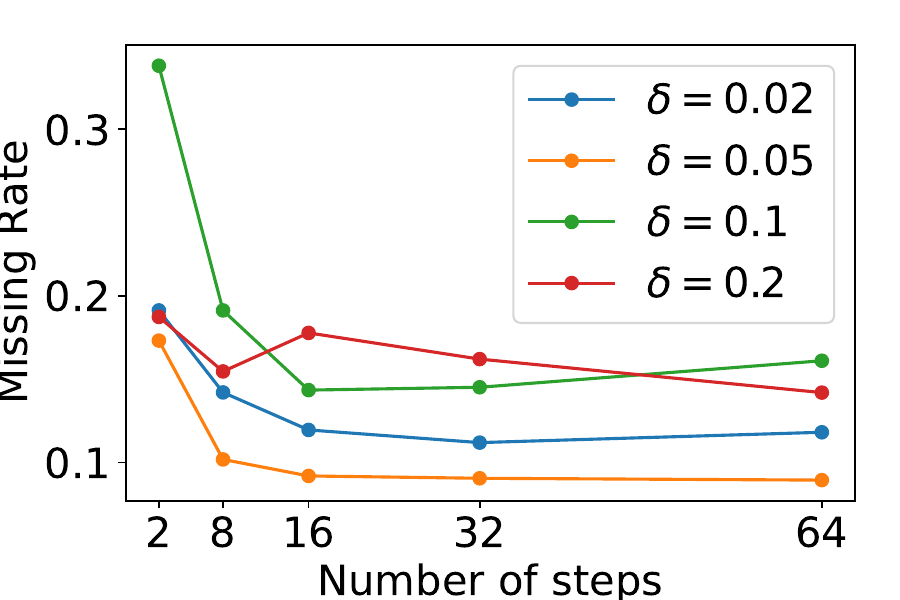} \\
	{\scriptsize \hspace{20pt}(a)\hspace{130pt}(b)\hspace{130pt} (c)}
	\caption{Influence of hyperparameters. Performance comparison of energy-based models with different numbers of Langevin steps and Langevin step sizes is shown in each sub-figure. Each curve represents a model with a certain Langevin step size $\delta$. We set $\delta$=0.02, 0.05, 0.1 and 0.2. For each setting of $\delta$, we choose different numbers of steps $l$=2, 8,16, 32 and 64.
		Performances are measured by (a) average RMSE, (b) minimum RMSE, and (c) missing rate on the Massachusetts driving dataset.}\label{fig:EBM_ablation}
\end{figure*}

\begin{figure*}[h]
	\centering
	\includegraphics[width=0.26\linewidth]{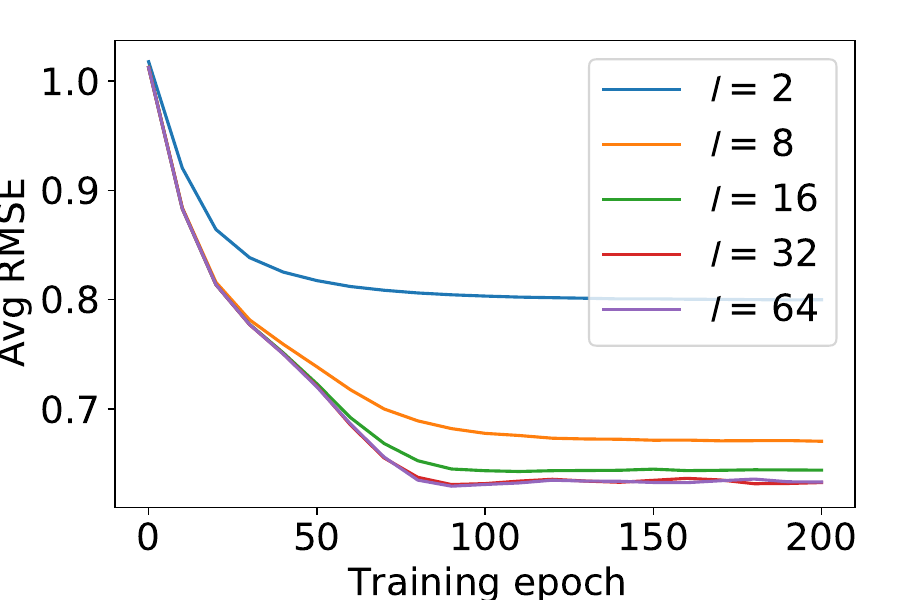}
	\includegraphics[width=0.26\linewidth]{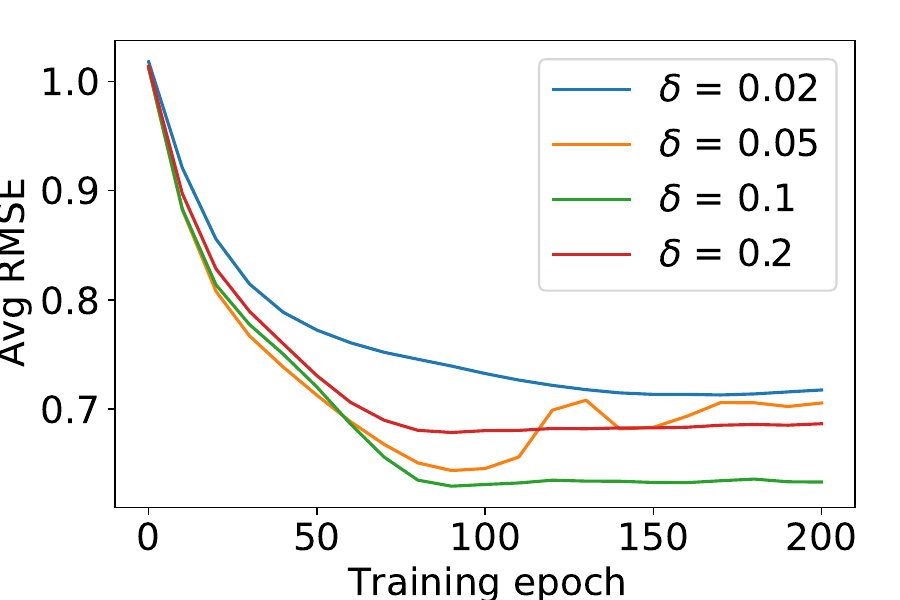}
	\includegraphics[width=0.26\linewidth]{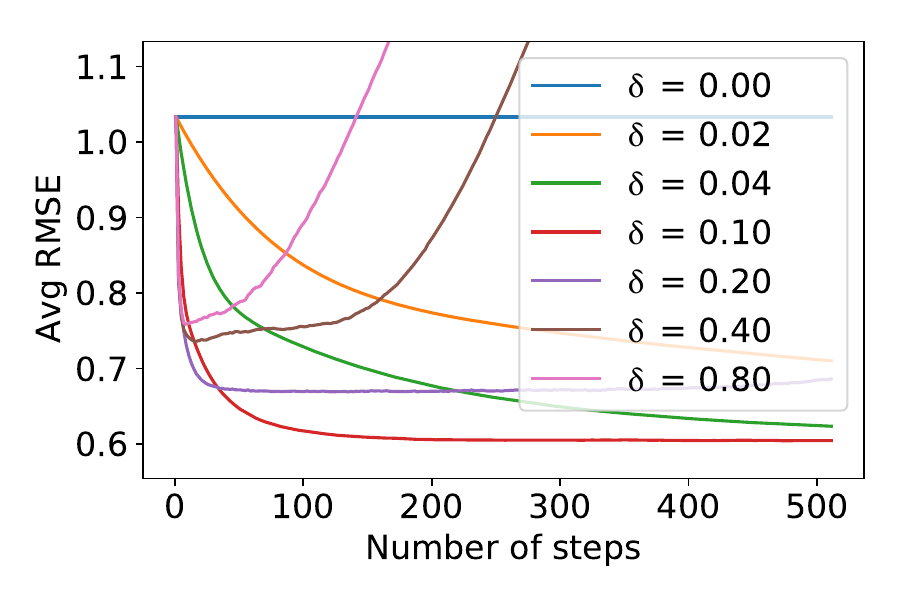} \\
	{\scriptsize \hspace{20pt}(a)\hspace{130pt}(b)\hspace{130pt} (c)}
	\caption{Influence of hyperparameters. (a) a line chart of testing average RMSEs over training epochs for different numbers of Langevin steps $l$ used in training. (b) a line chart of testing average RMSEs over training epochs for different Langevin step sizes $\delta$ used in training. (c) Influence of different numbers of Langevin steps $l$ and step sizes $\delta$ used in testing.}\label{fig:ablation}
\end{figure*}

We compare our method with the following baselines for multi-agent control.
\begin{itemize}
	\item Constant velocity: The simplest baseline with a constant velocity and a zero steering.
	\item The parameter sharing GAIL (PS-GAIL) ~\cite{bhattacharyya2018multi} \cite{simulation2019}: It extends the single-agent GAIL and the Parameter Sharing Trust Region Policy Optimization (PS-TRPO) \cite{gupta2017cooperative} to enable imitation learning in the multi-agent control~context.
	
\end{itemize}

We test our method on the NGSIM dataset. We use a linear cost function setting for each agent in this experiment. The maximum number of agents is 64.  Figure \ref{fig:multi_traj} shows two examples of the qualitative results. Each row is one example. The rows from first to fifth show the positions of all vehicles in the scene as dots at different timesteps respectively, along with the predicted trajectories (the green lines) and the ground truths (the red lines). Table \ref{tab:mul_ag} shows a comparison of performances between our method and the baselines in terms of RMSE.
Results show that our method can also work very well in the multi-agent control scenario.

\begin{table}[h]
	\centering
	\captionof{table}{Performance comparison in multi-agent control on the NGSIM dataset. Average RMSEs are reported.} \label{tab:mul_ag}
		\begin{center}
			\begin{tabular*}{\hsize-20pt}{@{\extracolsep{\fill}}l|cccc}
				\Xhline{2\arrayrulewidth}
				Method               & 1s    & 2s    & 3s    & 4s    \\
				\hline
				Constant Velocity           & 0.569 & 1.623 & 3.075 & 4.919 \\
				PS-GAIL & 0.602 & 1.874 & 3.144 & 4.962 \\
				ours (multi-agent)  & \textbf{0.365} & \textbf{0.644} & \textbf{1.229} & \textbf{2.262} \\
				\Xhline{2\arrayrulewidth}
			\end{tabular*} 		
		\end{center}
\end{table}

\subsection{Joint training with trajectory generator}

In this section, we follow Algorithm \ref{alg:3} to introduce a trajectory generator as a fast initializer for our Langevin sampler. In the experiment, we design $F_\alpha$ as a 4-layer MLP with output dimensions 64, 16, 8 and 2, respectively, at different layers. The activation function is ReLU for each hidden layer and Tanh for the final layer. The learning rate of the trajectory generator is set to be 0.005. We update the generator 5 times for each cooperative learning iteration. The rest of the setting remains the same as in the model with a linear cost function.  

Table \ref{tab:coop_res} compares the proposed joint training method with the following baselines in terms of average RMSE. The methods include (1) ``EBM w/o a generator'': the single EBM method without using a trajectory generator. (2) ``generator in joint training'': the trajectory generator trained with an EBM via the proposed cooperative training algorithm. We train both baseline methods (1) and (2) as well as our joint training framework (which we refer to as ``EBM with a generator'' in Table \ref{tab:coop_res} with different numbers of Langevin steps. Besides, we implement method (3), which is a single trajectory generator trained via maximum likelihood estimation with MCMC-based inference \cite{xie2019learningrep}.  

\begin{table}[h]
	\begin{center}
	\captionof{table}{Results of the joint training with trajectory generator on the Massachusetts driving dataset. (average RMSE)}\label{tab:coop_res}
		\begin{tabular*}{\hsize-2pt}{@{\extracolsep{\fill}}l|ccccc}
			\Xhline{2\arrayrulewidth}
			Number of steps & 2 & 8 & 16 & 32 & 64  \\	\hline
			EBM w/o a generator & 0.845 & 0.746 & 0.709 & 0.672 & 0.636 \\	 		
			generator in joint training & 0.956 & 0.835 & 0.844 & 0.845 & 0.854\\
			EBM with a generator & \textbf{0.804} & \textbf{0.672} & \textbf{0.649} & \textbf{0.638} & \textbf{0.633}\\
			\hline
			generator only &  \multicolumn{5}{c}{--------------------  0.911 --------------------  } \\
			\Xhline{2\arrayrulewidth}
		\end{tabular*} 	
	\end{center}
\end{table}

This comparison results show that a fast initializer can improve the performance even with less Langevin steps. For example, an EBM using 8 Langevin steps with a fast initializer is comparable with the one with a 32-step Langevin dynamics. Also, the method of ``generator in joint training'' performs better than the ``generator only'' setting because of the guidance of Langevin sampling of the EBM. 

\subsection{Training time and model size} We make a comparison of different methods in terms of computational cost and model size in the task of single-agent control on the Massachusetts driving dataset. We use a mini-batch of size 1,024 during training. For GAIL, the mini-batch size is 64. The total number of epochs is 40. Table \ref{tab:computation} lists the time consumption per training epoch and the number of model parameters for different settings on the Massachusetts driving dataset. The training time is recorded in a PC with a CPU i9-9900 and a GPU Tesla P100. As to the energy-based framework, a simple cost function design can lead to less computation time and less parameters. However, complex cost function can result in better performance in terms of RMSE. Overall, compared with the GAIL and IOC-Laplace baselines, our energy-based IOC methods are competitive.

\begin{table}[h]
	\begin{center}
		\caption{Comparison of computation cost and model size.}\label{tab:computation}
		\begin{tabular*}{\hsize-20pt}{@{\extracolsep{\fill}}ll|ll}
			\Xhline{2\arrayrulewidth}
			Method  & & time per epoch & $\sharp$ of parameters  \\
			\Xhline{1\arrayrulewidth}
			\multirow{4}{*}{EBM-IOC} & Linear & $\sim$ 3 mins & 11 \\
			& MLP & $\sim$ 10 mins & 2817\\
			& CNN & $\sim$ 12 mins & 44017\\
			& Joint Training & $\sim$ 3 mins & 3790\\ \hline
			GAIL & & $\sim$ 10 mins & 5893\\
			IOC-Laplace &  & $\sim$ 1 min & 11\\
			\Xhline{2\arrayrulewidth}
		\end{tabular*}
	\end{center}
\end{table}

\subsection{Hyperparameter analysis for energy-based IOC models}
\label{sec:ablation}

\subsubsection{Influence of the number of Langevin steps and the Langevin step size in the single EBM framework} 

We firstly study the influence of different choices of some hyperparameters,
such as the number of Langevin steps $l$, and the step size $\delta$ of each Langevin step. Figure \ref{fig:EBM_ablation} depicts the performances of energy-based IOC models with different $\delta$ and $l$ on the Massachusetts driving dataset. Each curve is associated with a certain step size $\delta$ and shows the testing performances over different numbers of Langevin steps. The performances are measured by (a) average RMSE, (b) minimum RMSE and (c) missing rate. In our experiments, we draw 5 samples from the learned model for prediction.  Missing rate is the ratio of scenarios where none of all 5 sampled trajectories has an endpoint L2 error less than 1.0 meters. The three metrics are used in the sub-figures of Figure \ref{fig:EBM_ablation}, respectively. In general, with the same $\delta$, the model performance increases as the number of Langevin steps increases. However, the performance gains become smaller and smaller while using more Langevin steps. Using more Langevin steps will also increase the computational time of sampling. We use $l=64$ to make a trade-off between performance and computational efficiency. We also choose $\delta=0.1$ for a trade-off among performances measured by different metrics.  

\begin{figure*}[h]
	\centering
	\includegraphics[width=0.26\linewidth]{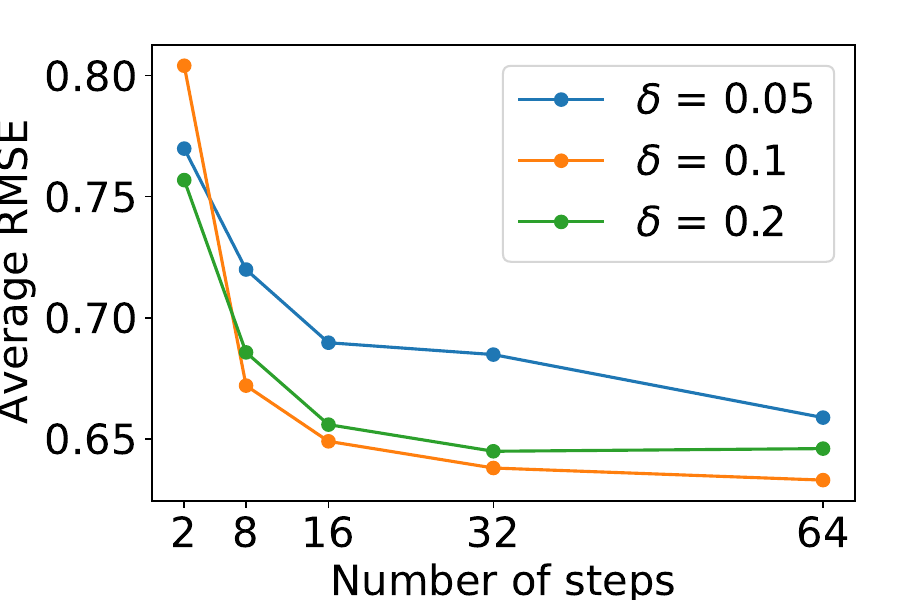}
	\includegraphics[width=0.26\linewidth]{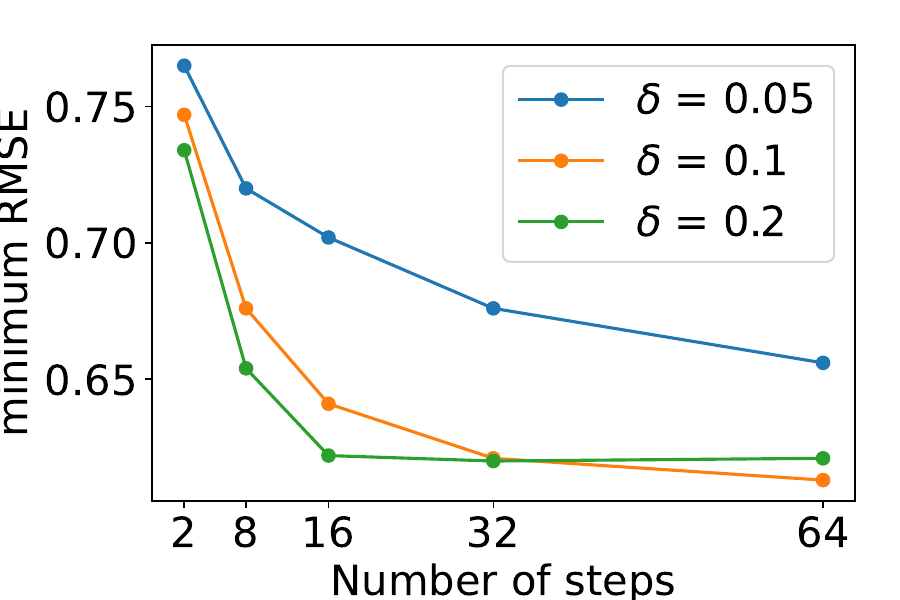}
	\includegraphics[width=0.26\linewidth]{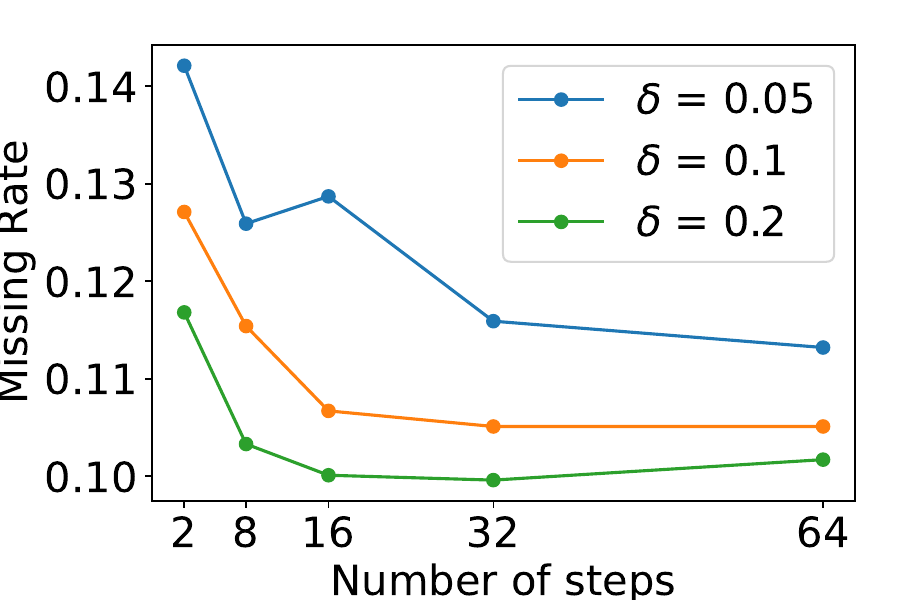} \\
	{\scriptsize \hspace{20pt}(a)\hspace{130pt}(b)\hspace{130pt} (c)}
	\caption{Influence of hyperparameters for the cooperative training framework. Performance comparison of frameworks using different numbers of Langevin steps and different Langevin step sizes is given in each sub-figure. Each curve represents a framework with a certain Langevin step size $\delta$. We set $\delta$=0.05, 0.1 and 0.2. For each setting of $\delta$, different numbers of Langevin steps are chosen, $l$=2, 8,16, 32 and 64. Performances are measured by three different metrics, which are (a) average RMSE, (b) minimum RMSE, and (c) missing rate.}\label{fig:coop_ablation}
\end{figure*}
\begin{figure*}[h]
	\centering
	\includegraphics[width=0.26\linewidth]{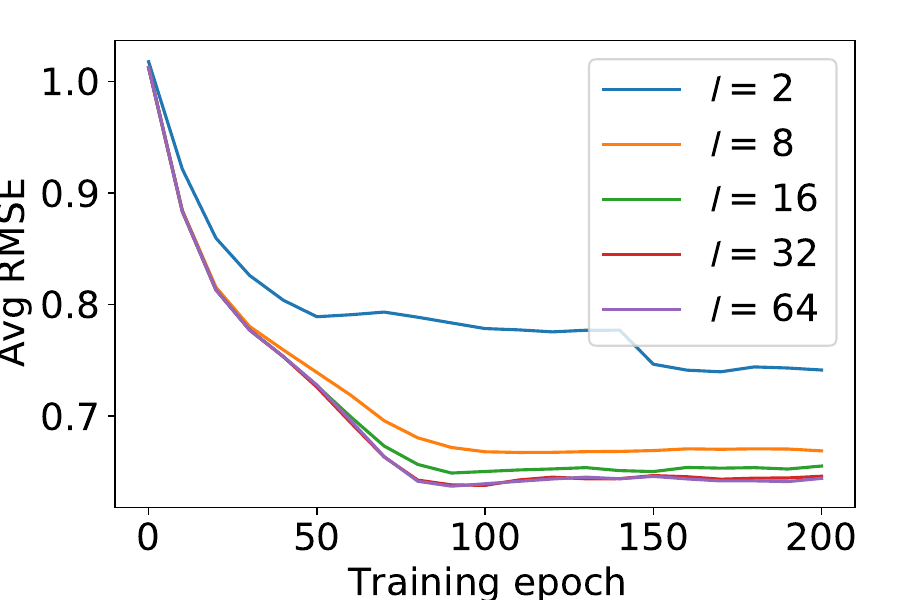}
	\includegraphics[width=0.26\linewidth]{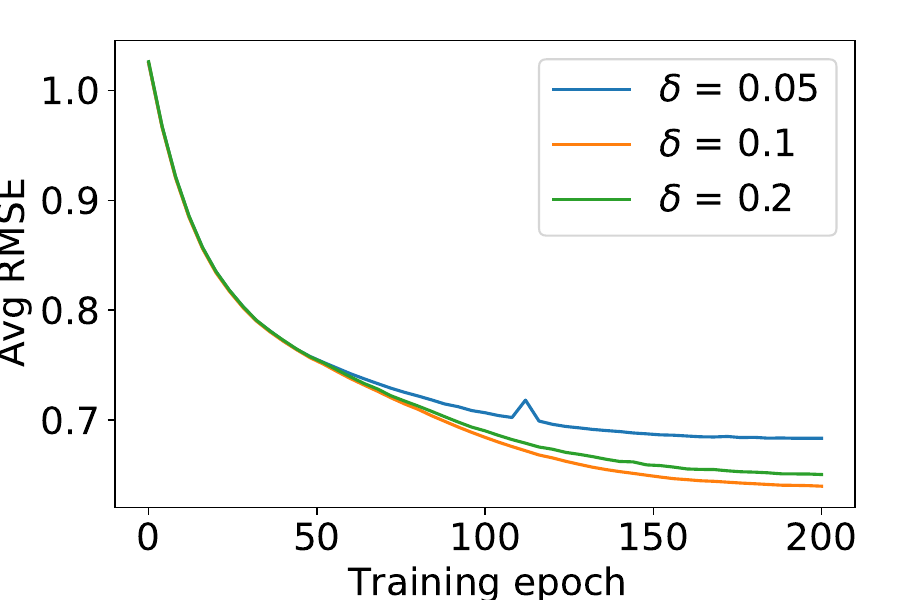}
	\includegraphics[width=0.26\linewidth]{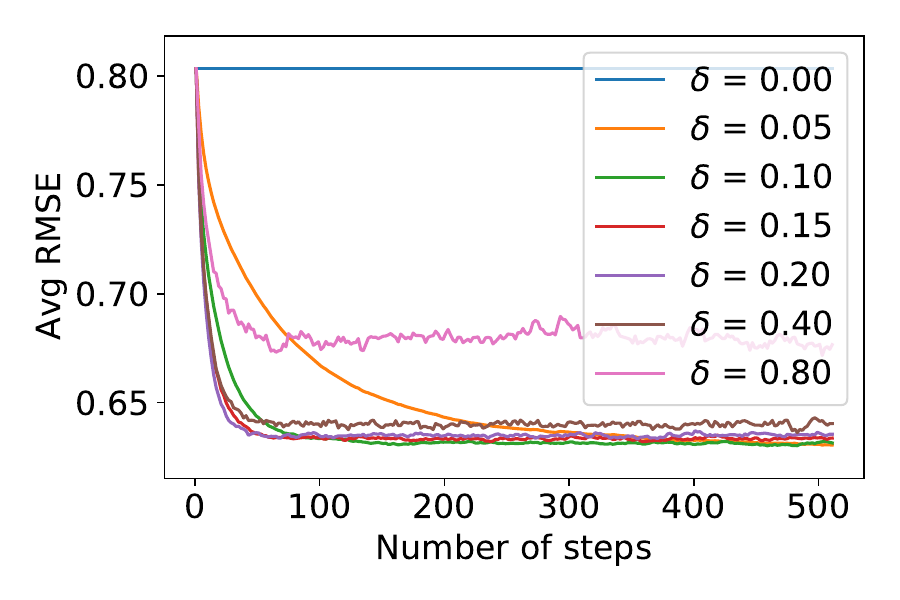} \\
	{\scriptsize \hspace{20pt}(a)\hspace{130pt}(b)\hspace{130pt} (c)}
	\caption{Influence of hyperparameters for the cooperative training framework. (a) Testing performances during the cooperative training with different Langevin steps $l$. (b) Testing performances during the cooperative training with different Langevin step sizes $\delta$. (c) Testing performances after training. Each curve shows testing performances over different numbers of Langevin steps used to sample trajectories in testing. Each curve is associated with a step size chosen in testing. The model is trained with the Langevin step size $\delta=0.2$ and the number of steps $l=32$.}\label{fig:EBM_testig_curve}
\end{figure*}

Figures \ref{fig:ablation}(a) and \ref{fig:ablation}(b) depict  training curves of the models with different $l$ and $\delta$, respectively. The models are trained on the Massachusetts driving dataset. Each curve reports the testing average RMSEs over training epochs. For testing, we use the same $l$ and $\delta$ as those in training. We observe that the learning is quite stable in the sense that the testing errors drop smoothly with an increasing number of training epochs. 

We also study, given a trained model, how different choices of $l$ and $\delta$ in testing can affect the performance of the model.  
Figure \ref{fig:ablation}(c) shows the average RMSEs of trajectories that are sampled from a learned model by using different numbers of Langevin steps $l$ and step sizes $\delta$. The model we use is with a linear cost function and trained with $l=64$ and $\delta=0.1$. We observe that: in the testing stage, using Langevin step sizes smaller than that in the training stage may take more Langevin steps to converge, while using larger ones may lead to a non-convergence issue. Thus, we suggest using the same $l$ and $\delta$ in both training and testing stages for optimal performance.

\subsubsection{Influence of the number of Langevin steps and the Langevin step size in the cooperative training framework}

We hence study the influence of different choices of the number of Langevin steps $l$ and the Langevin step size $\delta$ in the cooperative training framework. Figure \ref{fig:coop_ablation} depicts the performances of cooperative training frameworks with different $\delta$ and $l$ on the Massachusetts driving dataset. The performances shown in Figures \ref{fig:coop_ablation}(a), \ref{fig:coop_ablation}(b), and \ref{fig:coop_ablation}(c) are measured by average RMSE, minimum RMSE and missing rate, respectively. Each curve corresponds to a framework with a certain step size $\delta$ and shows the testing performances over different numbers of Langevin steps. Fixing $\delta$, the model performance increases as the number of Langevin steps increases. We use $l=32$ and $\delta=0.1$ to make a trade-off between performance and computational efficiency.

Figures \ref{fig:EBM_testig_curve}(a) and \ref{fig:EBM_testig_curve}(b) shows the learning curves of the cooperative training with different numbers of Langevin steps and different step sizes, respectively. Each learning curve shows testing average RMSE over different of training epochs. We observe that the testing average RMSE decreases smoothly as the number of training epochs increases. We also study how different $l$ and $\delta$ chosen in testing affect the performance of a learned energy-based IOC model. We first train an energy-based model with $\delta=0.2$ and $l=32$, and use the learned model in testing with varying Langevin step size $\delta$ and number of Langevin steps $l$. Figure \ref{fig:EBM_testig_curve}(c) depicts the influences of varying $\delta$ and $l$ in testing. We observe that given a learned cost function, Langevin sampling with a smaller step size and a larger number of Langevin steps may allow the model to generate better trajectories.

\section{Conclusion} \label{sec:con}

This paper studies the fundamental problem of learning the cost function from expert demonstrations for continuous optimal control. We study this problem in the framework of the energy-based model, and propose a sampling-based method and optimization-based modification to learn the cost function. Unlike the previous method for continuous inverse optimal control \cite{levine2012continuous}, we learn the model by maximum likelihood using Langevin sampling, without resorting to Laplace approximation. This is a possible reason for improvement over the previous method. Langevin sampling in general also has the potential to avoid sub-optimal modes. Moreover, we propose to train the energy-based model with a trajectory generator as a fast initializer to improve the learning efficiency. The experiments show that our method is generally applicable, and can learn non-linear and non-Markovian cost functions. 

In our future work, we shall explore other MCMC sampling or optimal control algorithms. We shall also experiment with recruiting a flow-based model \cite{DinhKB14,DinhSB17,KingmaD18} as a learned approximate sampler to amortize the MCMC sampling. We shall also adapt our model to the scenario where the human drivers may be sub-optimal, and some human judges may assign scores to the trajectories of some of the drivers. 

\ifCLASSOPTIONcaptionsoff
  \newpage
\fi

\ifCLASSOPTIONcompsoc
  % The Computer Society usually uses the plural form
  \section*{Acknowledgments}
\else
  % regular IEEE prefers the singular form
  \section*{Acknowledgment}
\fi

The work is supported by NSF DMS-2015577, DARPA SIMPLEX N66001-15-C-4035, ONR MURI N00014-16-1-2007, DARPA ARO W911NF-16-1-0579, DARPA N66001-17-2-4029, and XSEDE grant CIS210052.

\bibliographystyle{IEEEtran}
\bibliography{reference}

\begin{IEEEbiography}[{\includegraphics[width=1in,height=1.25in,clip,keepaspectratio]{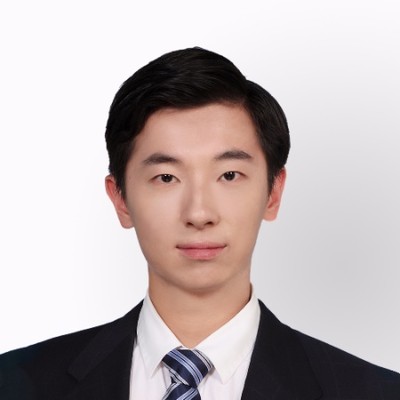}}]{Yifei Xu}
received his Ph.D. degree in statistics from University of California, Los Angeles (UCLA) in 2022. He received his B.E. degree in computer science at Shanghai Jiao Tong University. His research interests focus on genertive model, reinforcement learning and computer vision. 
\end{IEEEbiography}

\begin{IEEEbiography}[{\includegraphics[width=1in,height=1.25in,clip,keepaspectratio]{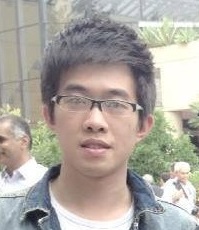}}]{Jianwen Xie}
received his Ph.D. degree in statistics from University of California, Los Angeles (UCLA) in 2016. He is currently a senior research scientist at Cognitive Computing Lab, Baidu Research, USA. Before joining Baidu, he was a senior research scientist at Hikvision Research Institute USA from 2017 to 2020, and a staff research associate and postdoctoral researcher in the Center for Vision, Cognition, Learning, and Autonomy (VCLA) at UCLA from 2016 to 2017. His research interests focus on generative modeling and unsupervised learning.
\end{IEEEbiography}
\begin{IEEEbiography}[{\includegraphics[width=1in,height=1.25in,clip,keepaspectratio]{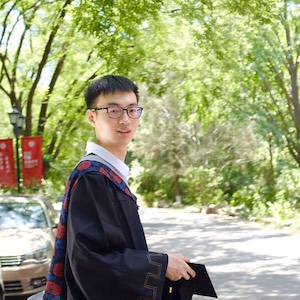}}]{Tianyang Zhao}
is currently a Ph.D. candidate in the Center for Vision, Cognition, Learning and Autonomy at the University of California, Los Angeles (UCLA). He received his B.E. degree in computer science at Peking University. His research interests lie in generative model and representation learning.
\end{IEEEbiography}
\begin{IEEEbiography}[{\includegraphics[width=1in,height=1.25in,clip,keepaspectratio]{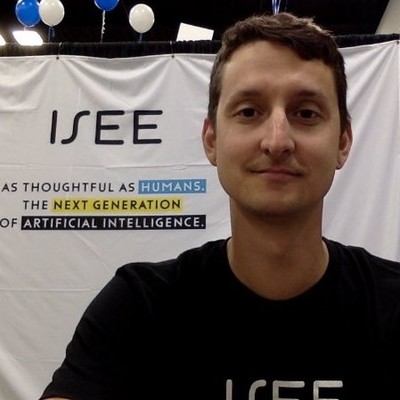}}]{Chris Baker}
received his Ph.D. degree in cognitive science from Massachusetts Institude of Technology (MIT) in 2012. He is currently a co-founder and chief scientist at iSee Inc. 
\end{IEEEbiography}
\begin{IEEEbiography}[{\includegraphics[width=1in,height=1.25in,clip,keepaspectratio]{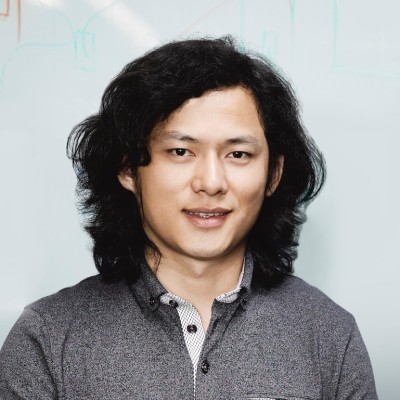}}]{Yibiao Zhao}
received his Ph.D. degree in statistics from University of California, Los Angeles (UCLA) in 2015. He is currently a co-founder and CEO at iSee Inc. 
\end{IEEEbiography}
\begin{IEEEbiography}[{\includegraphics[width=1in,height=1.25in,clip,keepaspectratio]{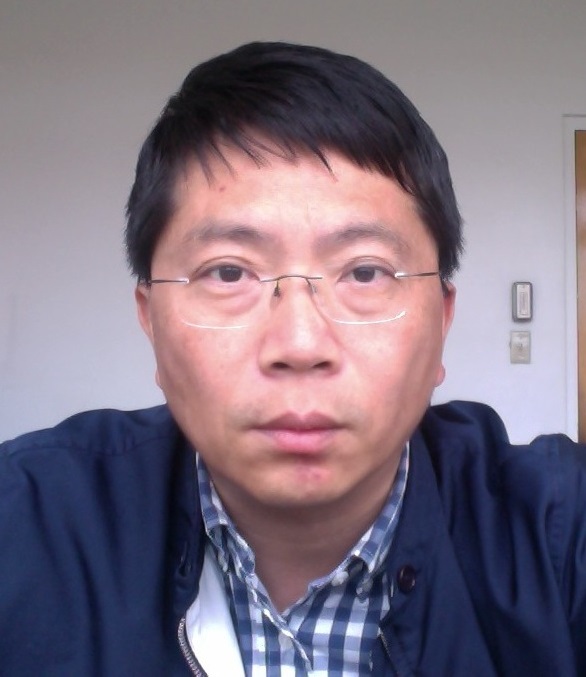}}]{Ying Nian Wu}
 received his Ph.D. degree in statistics from Harvard University in 1996. He was an assistant professor in the Department of Statistics, University of Michigan from 1997 to 1999. He joined University of California, Los Angeles (UCLA) in 1999, and is currently a professor in UCLA Department of Statistics. His research interests include generative models, representation learning, and computer vision. He received Honorable Mention for the David Marr Prize with S. C. Zhu et al. in 1999 and 2007 for generative modeling in computer vision.
\end{IEEEbiography}

\end{document}